%% file: main.tex
\documentclass{article}

\PassOptionsToPackage{sort&compress,numbers,super}{natbib}
\usepackage[sglblindworkshop, preprint]{neurips_2025}
\workshoptitle{SimBioChem 2025}

\usepackage[utf8]{inputenc} 
\usepackage[T1]{fontenc}    
\usepackage{hyperref}       
\usepackage{url}            
\usepackage{booktabs}       
\usepackage{amsfonts}       
\usepackage{nicefrac}       
\usepackage{microtype}      
\usepackage{xcolor}         
\usepackage{booktabs}
\usepackage{graphicx}
\usepackage{tabularx}
\usepackage{svg}
\usepackage{subcaption}
\usepackage{parskip}
\usepackage{nameref}
\usepackage[dvipsnames]{xcolor}
\usepackage{float}
\usepackage{listings}
\usepackage{xcolor}
\usepackage{natbib}

\newcommand{\agentname}{DynaMate}

\title{\agentname~: An Autonomous Agent for Protein-Ligand Molecular Dynamics Simulations}

\author{
  Salomé Guilbert\textsuperscript{1,\dag}, Cassandra Masschelein\textsuperscript{1,\dag}, Jeremy Goumaz \textsuperscript{1}, Bohdan Naida\textsuperscript{1},\\ \textbf{Philippe Schwaller\textsuperscript{1,2}} \\
  \textsuperscript{1}\'Ecole Polytechnique F\'{e}d\'{e}rale de Lausanne (EPFL) \\
  \textsuperscript{2}National Centre of Competence in Research (NCCR) Catalysis \\
  \texttt{\{salome.guilbert,cassandra.masschelein,philippe.schwaller\}@epfl.ch} \\
}

\begin{document}

\maketitle
\renewcommand{\thefootnote}{\fnsymbol{footnote}}
\footnotetext[2]{These authors contributed equally.} 
\renewcommand{\thefootnote}{\arabic{footnote}}

\begin{abstract}
  
  Force field-based molecular dynamics (MD) simulations are indispensable for probing the structure, dynamics, and functions of biomolecular systems, including proteins and protein–ligand complexes. Despite their broad utility in drug discovery and protein engineering, the technical complexity of MD setup, encompassing parameterization, input preparation, and software configuration, remains a major barrier for widespread and efficient usage. Agentic LLMs have demonstrated their capacity to autonomously execute multi-step scientific processes, and to date, they have not successfully been used to automate protein-ligand MD workflows. Here, we present~\agentname, a modular multi-agent framework that autonomously designs and executes complete MD workflows for both protein and protein–ligand systems, and offers free energy binding affinity calculations with the MM/PB(GB)SA method. The framework integrates dynamic tool use, web search, PaperQA, and a self-correcting behavior. \agentname~comprises three specialized modules, interacting to plan the experiment, perform the simulation, and analyze the results. We evaluated its performance across twelve benchmark systems of varying complexity, assessing success rate, efficiency, and adaptability. \agentname~reliably performed full MD simulations, corrected runtime errors through iterative reasoning, and produced meaningful analyses of protein–ligand interactions. This automated framework paves the way toward standardized, scalable, and time-efficient molecular modeling pipelines for future biomolecular and drug design applications.
\end{abstract}

\section{Introduction}

Molecular dynamics (MD) simulations are a cornerstone of computational chemistry and biophysics, enabling atomistic modeling of molecular interactions and conformational changes over time~\cite{hollingsworth2018molecular,karplus2002molecular,schlick2021biomolecular,schlick2011biomolecular}. 
Specifically, force field-based MD simulations provide crucial insights into the structure, stability, and function of biomolecules such as proteins, nucleic acids, and membranes, and have become indispensable tools in areas such as drug design and protein engineering. Further, conformational sampling of protein-ligand complexes, integrated in MD simulations, can be used to approximate binding free energies using the MM/PB(GB)SA method; molecular mechanics (MM) with Poisson–Boltzmann (PB)~\cite{davis1990electrostatics,honig1995classical} or generalized Born (GB)~\cite{feig2004recent,onufriev2019generalized}, and surface-area calculations~\cite{eisenberg1986solvation}. Such methods are increasingly used to rank a large number of structurally similar or congeneric binding complexes, since they offer reasonable accuracies at a reduced computational cost, compared to rigorous free-energy methods~\cite{nandigrami2022computational}. 

In practice, MD simulations are typically performed using specialized engines such as GROMACS~\cite{abraham2015gromacs}, OpenMM~\cite{eastman2017openmm}, or Amber~\cite{case2023ambertools}. 
However, preparing a molecular system for simulation remains one of the most error-prone and labor-intensive steps in the workflow~\cite{lemkul2024introductory}. 
After retrieving a structure from a database such as the Protein Data Bank (PDB)~\cite{burley2025updated}, users must perform structure cleaning, capping, protonation, solvation, and ionization. 
These preprocessing steps are followed by parameter selection (e.g., choice of force field, temperature, thermostat, barostat) and multi-stage equilibration prior to production runs. 
While protein parameterization is relatively straightforward due to mature force fields, handling small organic molecules, such as ligands or cofactors, is substantially more complex. 
Ligand parameterization requires compatibility between the biomolecular force field and the ligand’s charge and topology models, and even minor mismatches can lead to unstable or nonphysical trajectories. 
Given these challenges, automation offers clear benefits in reducing setup time, improving reproducibility, and enabling large-scale exploration of molecular systems.
Several frameworks have sought to automate MD simulations, including CHAPERONg~\cite{yekeen2023chaperong} and PyAutoFEP~\cite{carvalho2021pyautofep}, which facilitate simulation setup and analysis or integrate free-energy perturbation (FEP) workflows~\cite{lenselink2016predicting,valdes2021gmx_mmpbsa}. 
However, these pipelines remain rigid and domain-specific: extending them to new systems or simulation engines often requires significant manual modification and expert knowledge. 

Recent advances in large language models (LLMs)~\cite{radford2018improving,vaswani2017attention,brown2020language} have introduced new possibilities for scientific automation. 
LLMs can interpret natural language instructions, reason about domain-specific constraints, and orchestrate external tools, offering a flexible control layer that can adapt to new workflows and dynamically recover from errors. The emergence of agentic LLMs~\cite{m2024augmenting,skarlinski2024language,shinn2023reflexion} has demonstrated their capacity to autonomously execute multi-step scientific processes, including early efforts in MD workflow automation through MDCrow~\cite{campbell2025mdcrow} and NAMD-Agent~\cite{chandrasekhar2025automating}.
Both systems successfully automate protein-only simulations using OpenMM or CHARMM-GUI respectively, and can handle basic preprocessing and simulation stages. Yet, they remain limited in scope, unable to process protein–ligand systems, perform adaptive recovery from simulation errors, or generalize across simulation platforms.

There have also been efforts on text-to-code generation and fine-tuning of language models for the task with MDAgent~\cite{shi2025fine} which uses LAMMPS~\cite{LAMMPS} and a fine-tuned LLM on an internally curated dataset, albeit for the purpose of obtaining material thermodynamic parameters.

In this work, we introduce~\agentname, a modular multi-agent framework that autonomously designs and executes complete MD workflows for both protein and protein–ligand systems, with the option to perform free energy binding affinity calculations with the MM/PB(GB)SA method~\cite{valdes2021gmx_mmpbsa}.
Unlike prior systems, \agentname~separates high-level reasoning (e.g., parameter planning, structure retrieval) from low-level execution (e.g., simulation setup, file handling). 
This design enables dynamic tool use, retrieval-augmented parameter selection, and self-correcting behavior during simulation runs. 
By combining domain-aware reasoning with multi-agent coordination~\cite{wang2023voyager,shinn2023reflexion}, \agentname~bridges natural-language goal specification and robust computational execution in molecular simulation.

Table~\ref{SI:framework_comparison} summarizes key differences between existing agentic MD frameworks in the biology domain and~\agentname. Our system uniquely supports protein–ligand systems, integrates retrieval from external databases (web search as well as literature search with PaperQA \cite{narayanan2024aviarytraininglanguageagents,skarlinski2024language,lala2023paperqa}), and implements adaptive tool selection and error correction, thus extending the capabilities of current MD automation pipelines.

\section{Results}
\subsection{System Architecture: \agentname}
\begin{figure}[htbp]
    \centering
    \includegraphics[width=1\linewidth]{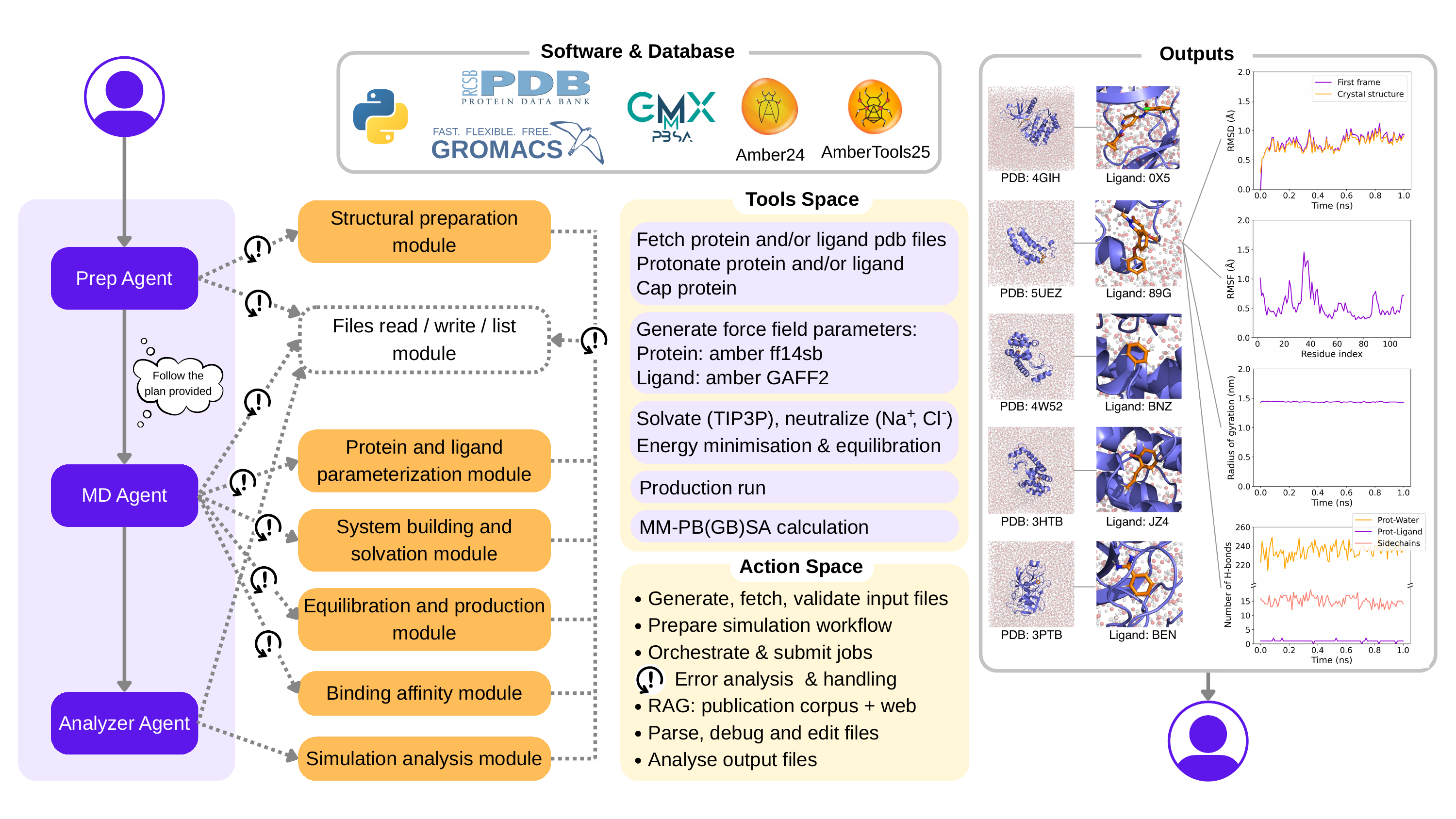}
    \caption{Overview of the framework. The Prep Agent constructs a context-aware simulation plan, the MD Agent executes it, and the Analyzer interprets the resulting trajectories. All agents use error-corrective reasoning. Representative output plots shown for the 5UEZ system: RMSD, RMSF, radius of gyration, and hydrogen-bond analysis.}
    \label{fig:main_workflow}
\end{figure}

\agentname~is a multi-agent architecture designed to automate the workflow of running MD for protein-ligand systems (Figure~\ref{fig:main_workflow}). The system is mainly comprised of a planner, a worker, and an analyser, and each agent is equipped with a set of tools and their descriptions. A plan is passed to the worker agent from the planner containing the system information such as the path to the PDB structure, the run directory, the name of the ligand if involved, and the desired simulation temperature and duration. A list of steps and corresponding descriptions are also passed to the worker agent. The plan is updated and reinforced in the agent-specific conversation history after each round of local decision-making and interactions in order to prevent confusion and loss of details. The agents also have access to their action traces and previous messages (or a summary). Action traces include whether tool calls failed and either the full error message or if the tool call proceeded successfully, a message summarizing the resulting output.

The core agent also has access to the run directory file structure containing the fetched (or provided) PDB as well as the \texttt{.mdp} (molecular dynamics parameters) files. This directory will grow as simulation files are created and the agent is able to edit, read, or create anything in the run directory. This, together with the provided tools allow the core LLM to reason, plan, act, and problem solve/critique in a complex setting where multi-step actions are required.

If the agent fails to complete the entire planned pipeline after 35 iterations (a hard limit imposed on the system) of tool calls and reflection, then the agent exits the execution. Upon exit, the agent provides a post analysis of the steps completed, any errors encountered during the pipeline, and any suggestions for further investigation or possible error correction. If the MD simulation ran successfully, this analysis includes the stability of the system's temperature, density and energy during equilibration, as well as an RMSD, RMSF, radius of gyration, and hydrogen bonding analysis. The agent also offers the ability to run binding free energy calculations in the event of a successful simulation and the presence of a ligand.

Throughout each workflow, the decisions of \agentname~are recorded, as well as the lapsed time for each agentic run, all files, and all potential error messages from the tools or software. This allows users to review decisions made by the agent as well as inspect the files created. Human experts can then verify the accuracy of the workflow and determine whether the appropriate error correction steps were taken. 

\subsection{Performance on protein-ligand complex MD simulations}
The performance and generalizability of \agentname~was evaluated across twelve distinct simulation setups. Five of these systems are well-established protein-ligand complexes that are widely used in the literature for benchmarking MD and free energy calculation methodologies (PDB IDs: 3HTB~\cite{boyce2009predicting}, 3PTB~\cite{marquart1983geometry}, 4GIH~\cite{liang2013lead}, 4W52~\cite{merski2015homologous} and 5UEZ~\cite{wang2017fragment}). Five systems did not contain a ligand, to test the agent's ability to adapt its workflow based on the system, and to simulate multi-chain proteins (PDB IDs: 1AKI~\cite{artymiuk1982structures}, 1FDH~\cite{frier1977structure}, 1J37~\cite{kim2003crystal}, 2CBA~\cite{haakansson1992structure} and 2VVB~\cite{sjoblom2009structural}). Finally, two systems were added with known errors or limitations: a system with an intentional atom name error in the input file (BRD4$\_$UNL) and a system with two ligands (PDB ID: 5KB6~\cite{Oliveira2016}). 
The selected systems (further detailed in Table~\ref{SI:systems_choice_rationale}) span diverse protein families and ligand chemotypes, providing a representative testbed for assessing the agent’s ability to set up, parameterize, and execute MD workflows across different biochemical contexts.

\begin{figure}[ht]
    \centering

    \begin{subfigure}[b]{0.9\linewidth}
        \centering
        \def\svgwidth{\textwidth}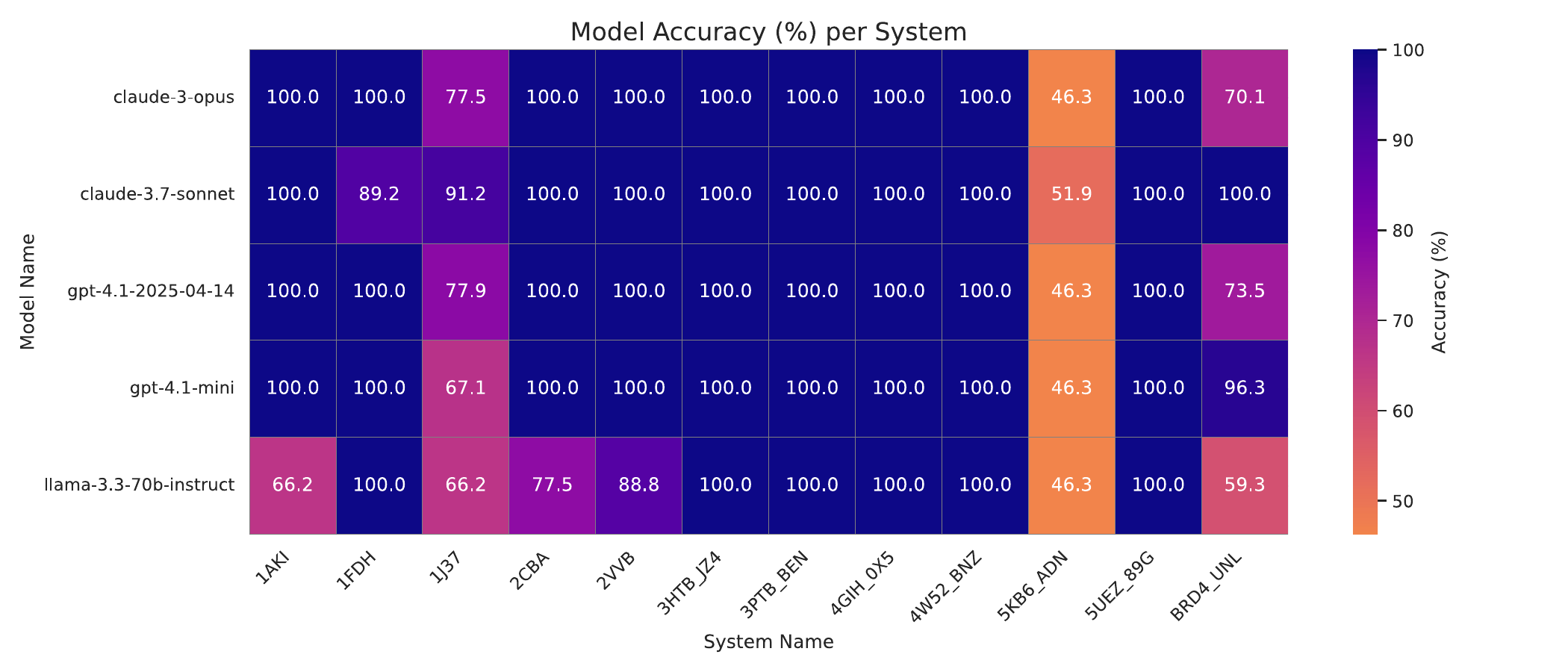
        \caption{Accuracy of the different LLMs for all tested protein and protein-ligand systems. Accuracy is defined for each essential step as the creation of the necessary (non-empty) files.}
        \label{fig:acc_heatmap}
    \end{subfigure}

    \vspace{1em} 

    \begin{subfigure}[b]{0.9\linewidth}
        \centering
        \def\svgwidth{\textwidth}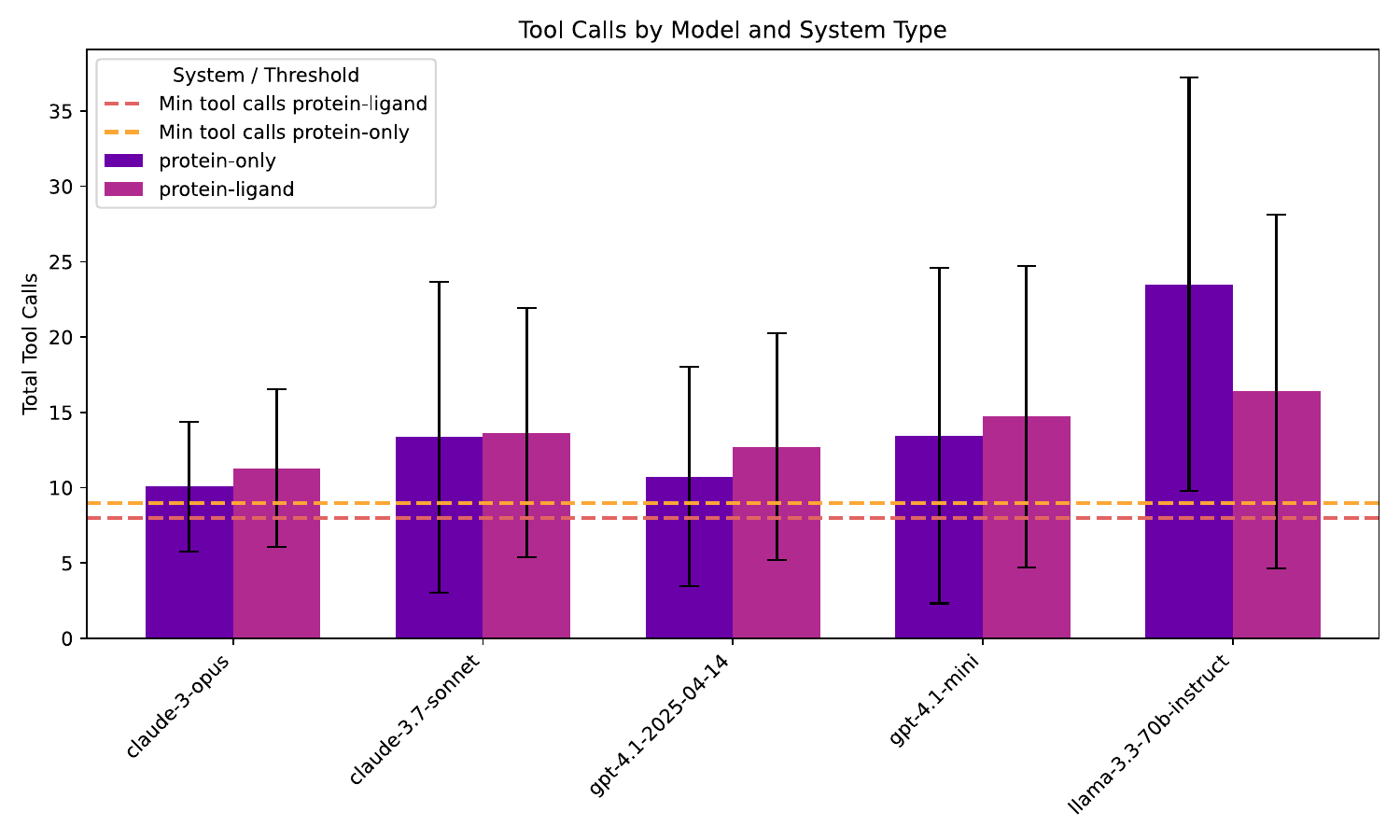
        \caption{Comparison of the numbers of tool calls by each model for both protein and protein-ligand systems. Here we aggregate all protein and all protein-ligand systems together.}
        \label{fig:tool_use_plot}
    \end{subfigure}

    \caption{The accuracy and efficiency of the different LLMs across the tested systems. Each system-LLM pair was ran three times and the average is reported. In (a) we report the accuracy for the entire pipeline as the average of the accuracies for each essential step. In (b) we denote the minimum number of required tool calls for each system type with a dashed line. This defines the number of essential tool calls required to successfully complete the task assuming no errors are encountered. The high variability in tool calls observed is a result of certain systems failing.}
\end{figure}

The agent's success was measured by its ability to correctly execute the tools required for the simulation to succeed, the creation of the necessary (non-empty) files, and the stability of the simulation. 
We evaluated the performance of \agentname~with five different LLMs detailed in Table~\ref{SI:tab_llm-overview}. Herein we will refer to the LLMs as Llama 3.3 \cite{touvron2023llama,dubey2024llama}, GPT-4.1 \cite{achiam2023gpt}, GPT-4.1 mini \cite{achiam2023gpt}, Claude 3 Opus \cite{anthropic2023model}, and Claude 3.7 Sonnet \cite{anthropic2023model}. Each system-LLM pair was tested 3 times in order to verify the robustness of our method. 
First, we see that \agentname~successfully prepared and performed a production MD run for all five protein--ligand systems, as evidenced by the 100$\%$ accuracy obtained for each agent on these systems (Figure~\ref{fig:acc_heatmap}). 
Proteins and ligands were correctly protonated at physiological pH, and the protein file was cleaned and capped. The complex systems were also correctly solvated, neutralized with ions, energy minimized, and equilibrated with 100 ps NVT and NPT simulations.
In addition, the output RMSD, RMSF, and radius of gyration plots were manually inspected and were consistent with equilibrated systems' outputs. Example plots for the 5UEZ$\_$89G system are shown in Figure~\ref{fig:main_workflow}. 

As mentioned, the 5KB6$\_$ADN and BRD4$\_$UNL systems contained known errors, and as a result the accuracies were lower for these two examples. 
Interestingly, the five agents varied in their approach to complete the workflow and achieved different degrees of accuracy.
In both test cases, Claude 3.7 Sonnet outperformed the other agents, reaching 100$\%$ accuracy for the BRD4$\_$UNL system across the three test runs. 
GPT-4.1 mini also performed very well for the BRD4$\_$UNL system, while the other three agents failed to solve the atom name error.
None of the agents were able to complete the workflow for the system with two ligands (5KB6$\_$ADN), highlighting limitations of \agentname~for this type of system.

\agentname~successfully adapted its workflow to simulate protein--only systems, including ones with multiple chains (1FDH). Llama 3.3 underperformed in each case, with a decrease in accuracy associated with the wrong use of tools (incorrect order or wrong variable called).
One system, 1J37, produced errors that no agent was able to completely solve, although once again, Claude 3.7 Sonnet outperformed the other agents.
Nevertheless, the different attempts to solve error messages demonstrated the utility of~\agentname, and are further discussed in section \ref{sec:correction_results}.


\subsection{Model Variability}
We evaluated the efficiency of \agentname~by measuring the number of tool calls over the minimum number of tool calls required to perform the simulation task. 
For instance, calling tools in the wrong order or using them with incorrect variables are causes of inefficiency losses. Additionally, input PDB or parameters files sometimes contain errors. Solving them with fewer calls of the file list / read / write tool results in efficiency gains. 
As seen in Figure~\ref{fig:tool_use_plot}, Claude 3 Opus and GPT-4.1 show strong performance in efficiently calling the right tools, minimizing the number of iterations needed to reach simulation completion. Claude 3.7 Sonnet and GPT-4.1 mini are slightly less efficient. As Claude 3.7 Sonnet was the best performing LLM in terms of accuracy, its less efficient score is associated with a larger number of tool calls in order to accurately solve pipeline errors. A further breakdown of the tool calling by system-type for each LLM can be found in \ref{SI:sec:extended_tool}.
Overall, these four LLMs show similar efficiency in performing both protein-only and protein-ligand simulations, highlighting the flexibility of~\agentname~to adapt to different system types, and the accuracy of the planner in generating simulation plans. 
Once again, Llama 3.3 is the least performing LLM, because of tool misuses and low error correction capabilities.

\subsection{\agentname~Error Correction Ability}\label{sec:correction_results}

\begin{figure}[ht]
    \centering
    \def\svgwidth{\textwidth}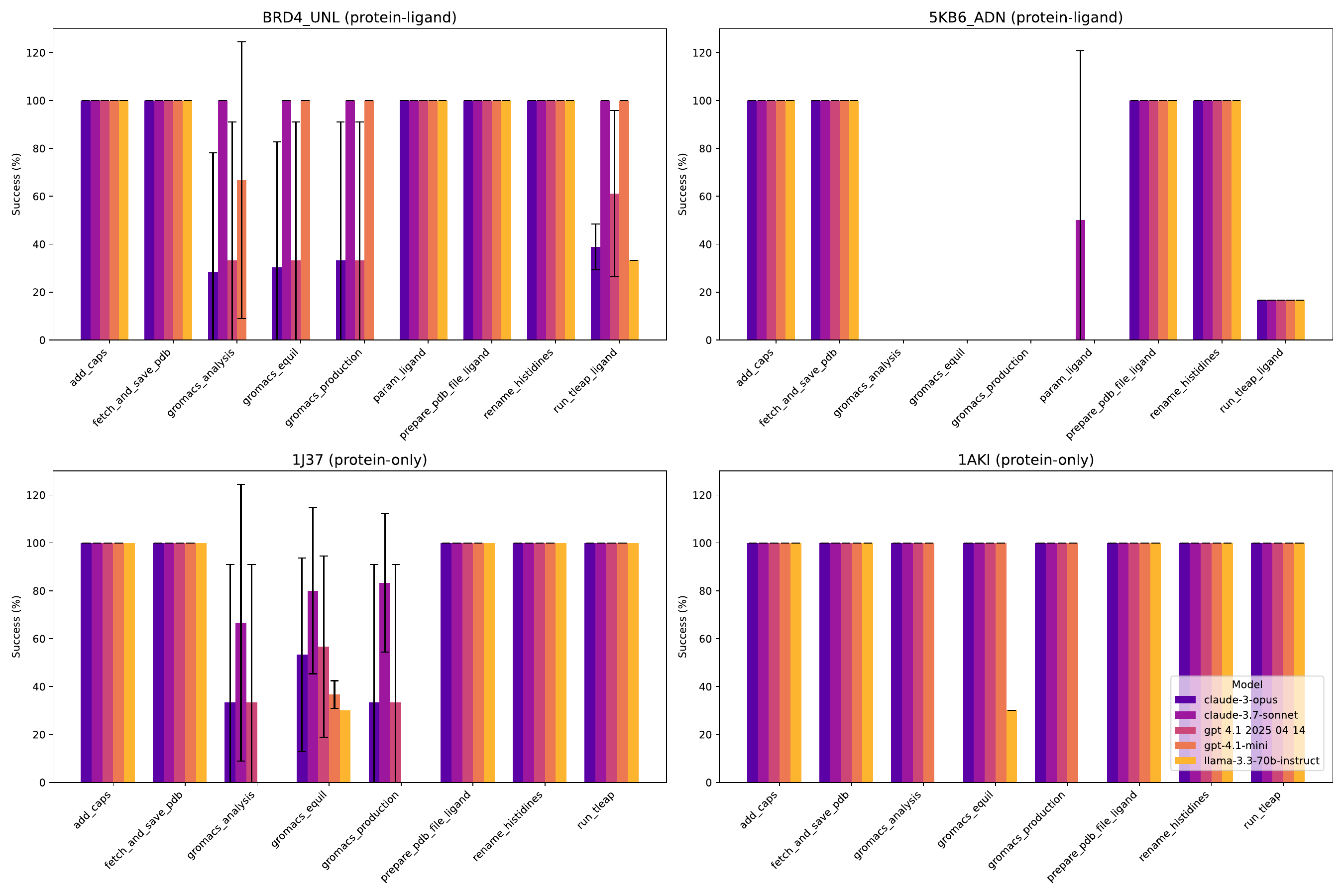
    \caption{The overall success of the pipeline can be broken down into the success rate of each required step. Here we show the breakdown of the success for each step for the different LLMs on the systems 1AKI, 1J37, 5KB6\_ADN, and BRD4\_UNL. We test each system three times to check for reproducibility.}
    \label{fig:problem_systems_combined}
\end{figure}

\textbf{Case study one: position-restraint errors} \agentname~encountered several issues during the generation of position-restraint files for the protein heavy atoms of the 1J37 system. Since this step is required for NVT and NPT equilibration phases, each agent experienced problems in the equilibration, production and analysis phases of GROMACS, as shown in Figure~\ref{fig:problem_systems_combined}. From Table~\ref{tab:position_restraints_error}, it appears that all agents correctly identified the cause of the errors and righfully directed their efforts towards position restraint files and GROMACS \texttt{.mdp} files (used for NVT and NPT equilibration). Recommendations given to the user also involved further investigation of these files. 
Claude 3.7 Sonnet was the only agent which manually created position restraint files itself, and completed the simulation. However, the generated files did not contain all the required heavy atoms, and were therefore not sufficient to perform proper equilibration of the system. In the case of Claude 3 Opus and GPT-4.1, the simulation was able to proceed after the reported corrections. These two agents completely removed the need for restraints as part of the equilibration, which also compromises proper equilibration of the protein system. A more granular breakdown of where in the pipeline the agents failed for the problematic systems can be found in \ref{SI:sec:extended_tool}.

\begin{table}[ht]
    \centering
    \small
    \setlength{\tabcolsep}{5pt}
    \caption{Examples of position-restraints error handling for the 1J37 system. Comparison of the behavior of multiple agents.}
    \label{tab:position_restraints_error}
    \vspace{0.3em}
    \begin{tabularx}{\linewidth}{
    >{\centering\arraybackslash}X 
    >{\centering\arraybackslash}X 
    p{0.3\linewidth} 
    p{0.3\linewidth} 
    >{\centering\arraybackslash}X}
        \toprule
        \textbf{Model} &
        \textbf{Error identified} &
        \textbf{Error correction attempts} &
        \textbf{Recommendations} & \textbf{Simulation complete} \\
        \midrule
        Claude 3 Opus & Yes &
        Removed position restraints in NVT and NPT equilibration runs &
        / & Yes \\
        Claude 3.7 Sonnet & Yes &
        Created custom position restraint file \texttt{posre.itp} applying a force constant of 1000 kJ/mol/nm$^2$ &
        Segmentation fault: check GROMACS compilation, reduce system size & Yes \\
        GPT-4.1 & Yes & Removed position restraints in NVT and NPT equilibration runs & Investigate index file generation, naming consistency, segmentation faults in mdrun & Yes \\
        GPT-4.1 mini & Yes & Tried to remove duplicate groups in index file causing selection problems and corrected posre.itp lines in topology file for each chain & Check topology file to ensure that position restraints for both protein chains are present, Index file should be cleaned to remove duplicates and inconsistent group names & No \\
        Llama 3.3 & Yes & Edited nvt.mdp and index.ndx file to include the correct group names & Investigate how the groups are defined in .mdp and index.ndx files & No \\
        \bottomrule
    \end{tabularx}
\end{table}

\textbf{Case study two: incorrect atom names}
For the BRD4$\_$UNL system, we tested~\agentname's ability to correct errors by purposely integrating an incorrect chlorine atom name in the input file, which resulted in an inconsistency with force field parameters. As a result, the Amber LEaP tool could not create the protein--ligand complex topology file without manual editing of the input PDB file. Interestingly, Claude 3.7 Sonnet and GPT-4.1 mini correctly solved the error in both of their three test runs. Both agents corrected the chlorine name in the input PDB file, which allowed the simulation to proceed successfully (Table \ref{tab:chlorine_error}). 
In two of the runs, Claude 3 Opus attempted to use Amber LEaP in different ways, but it was insufficient to allow the simulation to proceed. In the third test run, Claude 3 Opus completely removed the chlorine atom. Chlorine was then replaced by hydrogen during protonation, and the simulation pipeline completed. However, the structure of the ligand was changed, so the files generated are not satisfactory. 
Although two runs with GPT-4.1 were incomplete despite attempts to edit the input files and change atom names, one run succeeded with a different approach. The agent manually inserting correct GAFF values for \texttt{cl} into the ligand's force field parameters file. 
Finally, Llama 3.3 did not produce any successful simulations. The agent tried the tools multiple times, without trying to manually correct the errors in the input file.

\begin{table}[ht]
    \centering
    \small
    \setlength{\tabcolsep}{5pt}
    \caption{Examples of error handling for a wrong atom name in the input file for the BRD4$\_$UNL system. Comparison of the behavior of multiple agents. Some agents performed differently for the runs 1, 2 and 3 due to the stochastic nature of LLMs (we identify the run number with $r_1$, $r_2$ and $r_3$)}
    \label{tab:chlorine_error}
    \vspace{0.3em}
    \begin{tabularx}{\linewidth}{
    >{\centering\arraybackslash}X 
    >{\centering\arraybackslash}X 
    p{0.3\linewidth} 
    p{0.25\linewidth} 
    >{\centering\arraybackslash}X}
        \toprule
        \textbf{Model} &
        \textbf{Error identified} &
        \textbf{Error correction attempts} &
        \textbf{Recommendations} & \textbf{Simulation complete} \\
        \midrule
        Claude 3 Opus ($r_1$,$r_2$) & Yes &
        Attempt to modify parameterization step &
        Consult force field parameters files & No \\
        Claude 3 Opus ($r_3$) & Yes &
        Removed chlorine atom in ligand file &
        / & Yes \\
        Claude 3.7 Sonnet & Yes &
        Modified chlorine atom name in input PDB file &
        / & Yes \\
        GPT-4.1 ($r_1$,$r_3$) & Yes & Edit PDB and parameters files to change chlorine name, Regenerate ligand parameters after each edit and re-run tleap & Check atom names inconsistency, ensure ligand has a covalently--bound chlorine & No \\
        GPT-4.1 ($r_2$) & Yes & Manually inserted correct GAFF values for “cl” into the ligand’s frcmod file, Ensured atom names/types were consistent across ligand files & / & Yes \\
        GPT-4.1 mini & Yes & Modified chlorine atom name in input PDB file & / & Yes \\
        Llama 3.3 & Yes & Tried the tools multiple times & Check ligand and PDB input files, investigate log files generated & No \\
        \bottomrule
    \end{tabularx}
\end{table}

\textbf{Case study three: two ligands}
The next case study involved a protein with two ligands (PDB: 5KB6), to test cases where~\agentname~is not currently provided with the complete toolset. Several ligands present would require modifications of both ligands and complex parameter files, index files, as well as topology files. Although none of the agents successfully solved the issue, this case study demonstrated the feedback specificity of~\agentname, with each agent attempting multiple solutions to resolve the issue (Table \ref{tab:two_ligands_error}). Overall, both GPT-4.1 models and the Llama 3.3 model were less creative when attempting to solve the error, and many iterations solely consisted in trying the tools multiple times. On the other hand, Claude 3 Opus and 3.7 Sonnet demonstrated more diversity in their efforts to solve the problems. Claude agents made use of literature search, attempted to manually create Amber input files, and inspected ligand files multiple times. As a result, they acquired a better understanding of the error and provided clearer and more useful recommendations to the user. 

\begin{table}[ht]
    \centering
    \small
    \setlength{\tabcolsep}{5pt}
    \caption{Examples of error handling when multiple ligands are present for the 5KB6$\_$ADN. Comparison of the behavior of multiple agents.}
    \label{tab:two_ligands_error}
    \vspace{0.3em}
    \begin{tabularx}{\linewidth}{
    >{\centering\arraybackslash}X 
    >{\centering\arraybackslash}X 
    p{0.3\linewidth} 
    p{0.3\linewidth} 
    >{\centering\arraybackslash}X}
        \toprule
        \textbf{Model} &
        \textbf{Error identified} &
        \textbf{Error correction attempts} &
        \textbf{Recommendations} & \textbf{Simulation complete} \\
        \midrule
        Claude 3 Opus & Yes &
        Literature search to use pdb2gmx, Inspected the ligand file for any structural inconsistencies or formatting issues &
        Review the ligand structure and protonation state, provide the corrected ligand PDB file and rerun, standardize all atom names & No \\
        Claude 3.7 Sonnet & Yes &
        Created custom scripts to try to convert AMBER files to GROMACS format, Attempt to create a GROMACS topology directly using gmx pdb2gmx, Attempt to manually create parameters files &
        Investigate the ligand structure for irregularities that might cause antechamber to fail, Check duplicate atoms in the PDB file, Manually parameterize ligands & No \\
        GPT-4.1 & Yes & Tried the tools multiple times & Inspect and possibly regenerate the ligand (ADN) structure, Check if the ligand contains unusual or unsupported connectivity & No \\
        GPT-4.1 mini & Yes & Inspected and tried to fix the ligand PDB file, Attempt to run protein without ligand & Inspect ligand structure, Verify the ligand connectivity and atom types are compatible with antechamber & No \\
        Llama 3.3 & Yes & Internet search on tLEaP errors & Check Amber input parameters and ligand file, investigate outputs of parameterization tool & No \\
        \bottomrule
    \end{tabularx}
\end{table}

\subsection{Binding Free Energy with~\agentname}
\begin{figure}[htbp]
    \centering
    \includegraphics[width=1\linewidth]{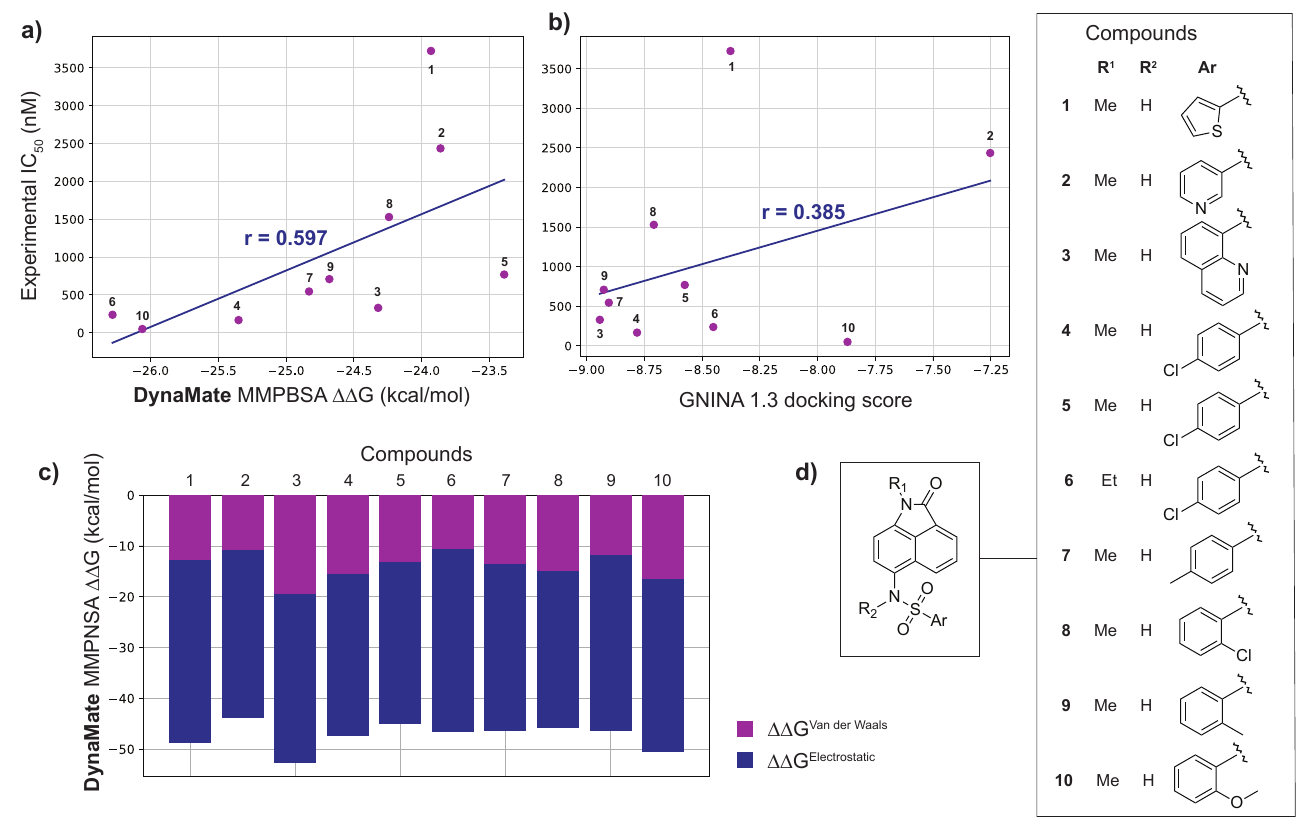}
    \caption{Comparison of the experimental $\mathrm{IC}_{50}$ binding affinity values with (a) ~\agentname\, MMPBSA $\Delta$$\Delta$G values and (b) GNINA 1.3 docking scores, for a set of BRD4 BD1 inhibitors~\cite{jiang2019discovery}. (c) Van der Waals and electrostatic contributions to the MMPBSA $\Delta$$\Delta$G~\agentname. (d) Scheme of the ten inhibitors tested for this study.}
    \label{fig:binding_affinity}
\end{figure}

Once a protein-ligand simulation is complete, \agentname~offers the option to perform MM/PB(GB)SA calculations to obtain ligand binding free energies, with the gmx\_MMPBSA tool~\cite{valdes2021gmx_mmpbsa,miller2012mmpbsa}.
The integration of this binding affinity tool was tested against a set of inhibitor molecules for the bromodomain 1 of Bromodomain-containing protein 4 (BRD4 BD1).
The set of ten inhibitors (compounds \textbf{16} to \textbf{25} first docked into BRD4 BD1 (PDB: 6JJ3~\cite{jiang2019discovery}) were also studied using the software GNINA 1.3~\cite{mcnutt2025gnina}.
\agentname~then autonomously performed 100 ns MD simulations of the complex structures and determined a binding free energy value with the PB solvation model in each case. 
The binding affinity values obtained with \agentname~and the GNINA docking scores were then compared with experimental $\mathrm{IC}_{50}$ values~\cite{jiang2019discovery}. 
As seen in Figure \ref{fig:binding_affinity}, $\Delta\Delta$G values given by \agentname's MMPBSA calculations are in reasonable agreement with experimental $\mathrm{IC}_{50}$ values (r=0.597). 
The MMPBSA method also offers valuable insights in addition to binding scores given by docking softwares. 
Not only does experimental evidence correlate better with MMPBSA values (r=0.597) than with GNINA docking scores (r=0.385) in our test case, but the MMPBSA method also provided more information about energetic components of binding. Indeed, van der Waals and electrostatic energetic contributions help rationalize the difference in binding between the ten inhibitors investigated. For instance, the most potent inhibitor (compound \textbf{10} from Figure \ref{fig:binding_affinity}) seems to bind more strongly than compound \textbf{1} (least potent inhibitor) mainly due to an increase in van der Waals interactions, while its stronger binding with respect to compound \textbf{2} seems to be linked to stronger electrostatic interactions. 


\section{Discussion}
Agentic systems like \agentname~illustrate the feasibility of ``AI scientists''~\cite{skarlinski2024language,wang2023scientific} capable of performing full experimental workflows in computational chemistry. Unlike rigid automation scripts that follow a deterministic workflow, our multi-agent architecture generalizes across simulation goals and tools by reasoning about both physical constraints and software syntax. In this section we will go over the three core-strengths of \agentname~and outline the current limitations and future directions.

\textbf{Error catching and self-diagnosing} Runtime errors can be challenging and require a significant amount of time by the user for debugging. Offsetting (some of) this responsibility to an agent can help to streamline scientific discovery. We observed the error correction capabilities through three case studies. 

First, we tested the agents' ability to deal with position restraints implementation in GROMACS by using a problematic system, 1J37. Imposing position restraints on the protein and ligand's heavy atoms is a crucial step during equilibration, and non-trivial to implement in MD softwares. Specifically, creating multiple position restraints file for each protein chain and for ligands, while referencing them in the correct position in the topology file is often a source of error. The tools provided to \agentname~led to several issues: the position restraint files for the two protein chains were not properly created, referenced in the topology file, and additional complications arose from inconsistent group names in the index files, which required manual editing. In this case, resolving the error involved multiple individual steps across several files, making it particularly challenging. Unsurprisingly, none of the agents fully resolved the issue (Table ~\ref{tab:position_restraints_error}). Nevertheless, each agent analyzed the error correctly and attempted partial solutions, illustrating the range of strategies that can be applied when addressing complex workflow errors. 
This case study also highlighted a key advantage of~\agentname: even when a system fails, the user receives a clear explanation of the underlying cause along with actionable recommendations for correction. The user is guided towards specific files with accurate keywords to look for, as all agents correctly identified problems in \texttt{.mdp}, index and topology files.

Case study two (chlorine atom name) clearly demonstrated a successful case of error catching and correction by~\agentname. All five agents tested correctly identified the error, and three of them correctly solved it, allowing the simulation to complete. This example also highlights that some errors have several working solutions; in this case, either editing the input PDB file or directly imputing the right parameters for the wrong atom name in the topology file. 

In the third case study, none of the agents were able to adapt the provided tools to accept a system with more than one ligand. This example demonstrates one limitation of~\agentname: it is not able to completely adapt its workflow when the modification of multiple files, in parallel, is required. However, the user was correctly informed that errors were linked with the ligand coordinate file and that with its available tools,~\agentname~was not able to construct appropriate parameters files.

Overall, even in cases where~\agentname~is not able to complete the simulation workflow, its subsequent attempts to deal with error messages allow it to gain a better understanding of the system and potential problems. As a result, it provides clearer and more useful recommendations for the user and indicates which files in particular to look into. This feature extends the use of~\agentname. A user that wants to investigate a known problematic system can receive advice regarding the errors in the parameterization, equilibration or production run.

\textbf{Literature search} An important component of~\agentname~is its ability to ground decisions in relevant scientific context. The combination of using PaperQA with a curated corpus as well as the web search tool enables the agent to quickly retrieve protocol-level information, recommended parameter choices, and community-reported solutions to common issues. We found that this hybrid system was particularly effective in cases where multiple plausible simulation strategies exist, such as determining protonation states, selecting force-field parameters, or interpreting ambiguous error messages. In these situations,~\agentname~was able to justify its choices by referring directly to published methods or authoritative documentation, reducing the burden on the user to manually verify each step. Although retrieval does not guarantee correctness, it substantially improves the transparency and traceability of the agent’s reasoning, and helps ensure that suggested workflows remain aligned with established practices. As autonomous systems begin to take on more complex scientific tasks, the ability to rapidly connect actions to citable sources becomes an essential safeguard, and our results suggest that~\agentname~benefits significantly from this retrieval layer.

\textbf{Simulation output analysis}
The manual inspection of output files confirmed the successful MD simulations of most protein and protein--ligand systems with 100$\%$ accuracy, shown in Figure~\ref{fig:acc_heatmap}. 
We noted one exception: GPT-4.1 removed a chlorine atom in the ligand of BRD4$\_$UNL, for one of the three runs performed with this agent (Table~\ref{tab:chlorine_error}). 
GPT-4.1 clearly stated its action in the summary and analysis given at the end of the run, which facilitates the user's ability to catch such undesirable behavior.


\textbf{Limitations}
The reliance on external LLM API's is a clear limitation with the current framework. This requires that users have the relevant API keys and available credits to run the model. There are also potential issues with rate limits and security concerns. We also do not include persistent memory for the agent, so each new run has no recollection of any prior runs. Incorporating long-term memory in addition to the working memory will be studied in future iterations of~\agentname. The clear underperformance of the Llama 3.3 agent, compared to Claude and GPT-4.1 ones, also highlights the current dependency of~\agentname's performance on the type of agent used. The agents tested demonstrated contrasting strengths, with respect to error analysis, error correction, literature search and simulation output analysis. 

In addition, the current toolset limits the range of systems that can be reliably simulated. For example, the tools do not explicitly support systems containing multiple ligands, and the agents were unable to overcome this limitation to generate complete parameters for such cases, as illustrated by the 5KB6$\_$ADN system. Expanding the current toolset to accommodate a broader set of biomolecular systems, such as membrane proteins, DNA and RNA, would increase the applicability of~\agentname. 
Some limitations also arise from unconventional atom nomenclature in input PDB files. Although the agents occasionally demonstrated successful error-correction behavior under these conditions (e.g., for the BRD4$\_$UNL system), non-standard atom names, missing residues, and incomplete coordinate files remain persistent sources of failure. While agent reasoning can help mitigate these issues, complementary efforts should focus on enhancing the robustness of the underlying tools to better handle such irregularities.

It is important to note that MD simulations of biological systems are highly case-specific. Although ~\agentname~uses generalized tools, has access to literature and to the internet, some system-specific parameters are at risk of being missed. In particular, the protonation of histidines, aspartic acid and glutamic acid residues can be difficult to determine. Knowing the pH at which a protein operates can help in such a decision, but still, the correct protonation of active site amino acid is not always clear; yet, these choices heavily influence simulation outputs. The same applies to ligands, for which the protonation has to be investigated in detail. Here, \agentname~assigns GAFF2 parameters to the ligand, but for reliable results, the user should check that they are correct for the small molecule of interest. Dihedrals and point charges can be further investigated to confirm that the simulation is as accurate as possible. 

Concerning the calculation of binding affinities, the limitations of the MM/PB(GB)SA method are well-known~\cite{roux2024editorial}. Approximations are introduced to lower the computational cost compared with more rigorous free-energy calculations that rely on explicit solvent all-atom MD simulations. In fact, MM/PB(GB)SA only integrates the conformational sampling from MD simulations in explicit solvent. The free energy itself is estimated from an implicit, continuum solvent model from PB, GB and SA calculations. These approximations lower the computational cost compared with more rigorous free-energy calculations in explicit solvent. Therefore, applying this computational method requires a critical understanding of the theoretical limitations and knowledge of the cases when it can be reliably used. For instance, such approximations generally fail to accurately represent changes in the conformational entropy of protein and ligand, as well as adjustments between the bound and unbound states. Such errors usually cancel out when ranking similar complexes, and MM/PB(GB)SA is mostly used in comparative studies.


Tools like~\agentname~lower the barrier to entry for running complex scientific workflows, thus making MD simulations more accessible to the broader scientific community. The streamlined usability of \agentname~enables more efficient incorporation of computational binding-affinity calculations into drug-optimization campaigns, and the modularity of the agentic architecture will allow the facile extension of possible experiments. The context-aware nature of our system makes it possible to suggest the most appropriate workflows for a series of new tasks. 
Future works will introduce diverse systems into the pipeline, such as DNA, membrane proteins, and multimers containing several ligands. There will also be integration of software specific documentation and on-demand retrieval of scientific papers for new systems.

\section{Methods}
\subsection{\agentname~Infrastructure}
Our LLM-based system was implemented with LiteLLM (v.1.79.3) including customized tool use, working memory, and error parsing and feedback. The models were accessed through OpenRouter with the online setting so that we could utilize the web search functionality. The specific details of the models used for the agents, Anthropic's Claude 3.7 Sonnet and Claude 3 Opus, OpenAI's GPT-4.1 and GPT-4.1 mini, as well as Llama 3.3 can be found in Table~\ref{SI:tab_llm-overview}, with model selection guided by a comparative evaluation of latency, cost, reasoning performance, and stability in tool-use–heavy workflows. A temperature setting of 0.1 was selected in order to ensure highly consistent, deterministic reasoning and minimize randomness of tool calls in planning multi-step actions.

Each agent is initialized with a personal context in the system prompt (as found in S\ref{SI: system_prompts}), and the custom plan from the planner agent is dynamically incorporated into the worker agents prompt for each run. After each tool call, the agent is provided with a summary of the completed tasks as well as the tasks remaining from the original plan. We also implement custom logic for software errors in order to route the context back to the LLM.

Finally, in order to minimize context overload and forgetting, we periodically summarize the entire chat history into a single assistant message, and replace earlier parts of the chat with it. For robustness, we include the last two assistant messages with custom logic such that tool calls and outputs are not broken in the message history.

\subsection{Agentic Workflow}
Our system,~\agentname, is designed as a multi-agent framework working in an environment of tools that autonomously prepares, executes, and analyzes MD simulations. 
The architecture extends previous agentic approaches such as MDCrow~\cite{campbell2025mdcrow} and NAMD-Agent~\cite{chandrasekhar2025automating} by introducing a dedicated planner, dynamic feedback with error correction, and by enabling simulations of protein--ligand complexes. 

The workflow is organized into three main components and is contained within a sandboxed environment:

\begin{enumerate}
    \item \textbf{Planner agent}: Extracts user intent and scientific requirements from natural language input, retrieves structural information, and constructs a context-aware simulation plan. 
    It identifies whether the system contains a ligand, decides on parameters such as simulation temperature and duration, and defines an ordered sequence of simulation subtasks for the system (preprocessing, equilibration, production, and analysis).
    The planner leverages retrieval-augmented reasoning~\cite{lewis2020retrieval} and tool invocation to retrieve the PDB file and curate an appropriate plan for the MD simulation.
    
    \item \textbf{Molecular dynamics agent}: Interprets the plan produced by the Planner agent and autonomously executes it through a loop of tool invocation → validation → reflection~\cite{shinn2023reflexion}.
    Each subtask is validated against the filesystem and output logs, allowing the agent to detect and repair common runtime and software errors by regenerating or modifying input files (e.g., correcting missing atom names or updating topology directives). 
    
    \item \textbf{Analyzer agent}: Post-processes MD trajectories and outputs descriptive analyses. The agent further provides qualitative summaries of stability, convergence, and ligand–protein interactions in natural language form.

    \item \textbf{Execution environment}: The agents operate within a sandboxed directory structure, ensuring safe read/write operations and reproducibility.
    Interactions with external LLMs are routed through LiteLLM, which acts as a unified interface for model invocation, request routing, and provider abstraction.
    The framework supports both AmberTools and GROMACS, expanding prior works restricted to OpenMM or CHARMM-GUI.
\end{enumerate}


A representative workflow is shown in Figure~\ref{fig:workflow_tools_protein_ligand}, illustrating the difference between a protein-only simulation and a protein-ligand complex. The planner dynamically selects additional steps for ligand parameterization and topology construction, automatically adjusting the downstream MD setup.


\subsection{\agentname~Toolset}
Our framework interacts with a structured set of domain-specific tools, defined through a formal schema that includes tool names, descriptions, and JSON input specifications. 
The tool schemes are loaded at runtime and the LLM selects actions through the tools and input arguments.
The major tool families include structural preparation, ligand parameterization, system building and solvation, equilibration and production, and web and data retrieval.

\textbf{PDB retrieval}: If not provided by the user, the PDB coordinate file is fetched from the Protein Data Bank (PDB)~\cite{burley2025updated}.

\textbf{Structural preparation}: The protein coordinate is extracted from the relevant PDB file. If a ligand is requested by the user, its coordinates are also extracted. The protein termini are capped with acetyl (ACE) and methionine (NME). Atom names in the protein and ligand are corrected to match force field parameters used in later steps.

\textbf{Ligand parameterization}: If a ligand is present, it is protonated using OpenBabel~\cite{o2011open} at pH 7.0, unless specified otherwise by the user or by the agent via literature search. The ligand is then parameterized with the antechamber programme of AMBER version 20~\cite{wang2006automatic,case2020amber2020}. The GAFF2 force field is applied, commonly used to parameterize small organic molecules~\cite{wang2004development}.

\textbf{Protein parameterization and system preparation}: The AMBER ff14sb force field is applied for the protein~\cite{maier2015ff14sb}. The system is solvated with the the TIP3P model for water~\cite{mark2001structure}, and neutralized with Na$^{+}$ and Cl$^{-}$ ions. 

\textbf{Equilibration and production}: The MD simulations are performed with the GROMACS version 2023 package~\cite{abraham2015gromacs}.
In order to simulate a realistic biochemical environment, periodic boundary conditions are applied. 
An integration time-step of 2 fs is employed for all simulations and covalent bonds with hydrogen are constrained using the LINCS algorithm~\cite{hess1997lincs}.
The Verlet cut-off scheme~\cite{verlet1967computer} is used throughout all steps with a force-switch modifier starting at 10 Å and a cut-off at 12 Å. 
The Particle-mesh Ewald (PME) method is used for long-range electrostatics and a cut-off of 12 Å was used for short-range
electrostatics~\cite{darden1993particle}.
As is standard practice to ensure low energy starting configurations, each system is initially energy minimized for 5000 steps of steepest descent or until
a maximum force of 1000 kJ.mol-1.nm-1 is reached on any atom. 
Following energy minimization, 100 ps equilibration-runs are performed in the NVT and NPT ensembles, maintaining the temperature at 300 K (or else if specified by the user or by the agent's literature search capability) and the pressure at 1 bar using the Berendsen thermostat and barostat, respectively~\cite{berendsen1984molecular}.
In the cases of simulations with a ligand, two temperature coupling groups ensure better temperature control on a system with one group consisting of the protein and the ligand, and the other of the solvent (water and ions).
Finally, NPT production runs are performed for various lengths.

\textbf{Analysis}: Trajectories are analyzed using GROMACS tools: gmx$\_$rmsd for root-mean-square deviation (RMSD) of the protein backbone (with respect to the first frame or the crystal structure), gmx$\_$gyrate for radius of gyration, gmx$\_$rmsf for root-mean-square fluctuation (RMSF) of protein residues, and gmx$\_$hbond for hydrogen bond quantification.

\textbf{Binding affinity calculations with the MM/PB(GB)SA method}: Binding free energies are computed using the gmx$\_$MMPBSA tool (version 1.4.3) with MMPBSA.py (version 16.0)~\cite{valdes2021gmx_mmpbsa,miller2012mmpbsa}.
The binding free energy is calculated as $\Delta G = \Delta H - T \Delta S$, where $\Delta H$ is the enthalpy change, $T$ is the temperature, and $\Delta S$ is the entropy change.

\textbf{Web and data retrieval}: web search or PaperQA \cite{narayanan2024aviarytraininglanguageagents,skarlinski2024language,lala2023paperqa} queries to determine experimental conditions such as recommended temperature or ligand information. Can also be used to diagnose common runtime errors.


This modular design allows for tool reuse and extension across multiple agents and systems, and facilitates transparent LLM–tool interactions.

\subsection{Error Correction and Feedback} A key feature of the framework is its iterative feedback mechanism, which is similar to the mechanism outlined in~\cite{zou2025agente,shinn2023reflexion}, enabling self-correction for failed or incomplete tool calls. \agentname~parses the errors from the failed tools which will include the error messages from the respective software programs such as GROMACS, AmberTools, or other general Python errors. Providing this rich context to the agent allows future steps to involve self-correction, such as editing input files with malformed structures, updating atom names to follow expected conventions, and running essential steps that might have been missed resulting in the creation of necessary files that were missing.

Each tool invocation is sandboxed, logged, and summarized into a compressed message context to maintain continuity between attempts.
If a subtask fails (ie. due to missing files, naming errors, or incompatible parameters) the agent analyzes the log output and proposes a corrected re-execution plan.

\subsection{Retrieval-Augmented Generation for Parameter Selection and Error Control}
To support literature-grounded decision-making during the design and execution of MD simulations, we integrated PaperQA \cite{narayanan2024aviarytraininglanguageagents,skarlinski2024language,lala2023paperqa}, a retrieval-augmented question-answering tool that operates over a curated corpus of 90 domain-relevant journal articles. These journal articles are specific to the systems tested in this study, and can be adapted to any other system of choice. PaperQA performs structured retrieval by embedding both user queries and full-text documents into a shared semantic space, followed by passage-level ranking and answer synthesis. Unlike generic RAG pipelines, PaperQA is optimized for scientific texts: it uses metadata-aware chunking, citation-preserving retrieval, and domain-specific answer constraints to ensure that all generated responses are explicitly grounded in referenced source passages.

Within our agentic workflow, PaperQA serves two key purposes. First, it enables the agent to search through relevant literature for appropriate system parameters and methodologies before launching MD runs, such as protonation states, whether the ligand works better at a specific pH or temperature, and whether the protein was crystallised under specific conditions that would require some modifications to the PDB file. Second, it provides real-time resources that can help understand errors in input systems that require correction. This substantially reduces human supervision, enabling autonomous workflows that remain aligned with established practices. Some examples of questions asked by the agent to PaperQA are displayed in the Table~\ref{table:seach_papers_questions} and some answers generated by PaperQA in Table~\ref{SI:seach_papers_answers}.

\begin{table}[ht]
    \centering
    \caption{Examples of questions asked by the agent to PaperQA as part of the {\ttfamily search\_papers} tool. The answers are obtained through a RAG system searching in a curated corpus of 90 domain-relevant journal articles.}
    \begin{tabular}{|l|}
        \hline
        \multicolumn{1}{|c|}{\textbf{Examples of agent questions to PaperQA (categorized)}} \\
        \hline\hline
        \textbf{MD simulation parameters selection} \\
        \hline
        Question: molecular dynamics simulation temperature for 3PTB protein \\
        Question: Why are MD simulations performed at room temperature? \\
        Question: short molecular dynamics simulation duration \\
        \hline\hline
        \textbf{Protonation states} \\
        \hline
        Question: Protonation states of histidine residues in protein simulations at pH 7 \\
        Question: protonation and charge states of ligand ADN at pH 7 for molecular dynamics simulations \\
        Question: protonation and atom types for ligand ADN in protein simulations at pH 7 \\
        Question: protonation state and parameterization of ligand ADN for Amber force field at pH 7 \\
        \hline\hline
        \textbf{Error solving and softwares} \\
        \hline
        Question: how to use gmx pdb2gmx to generate a gromacs topology from a pdb file \\
        Question: Why does ParmEd fail with a FileNotFoundError when using run\_tleap? \\
        Question: Common issues with GROMACS energy minimization infinite forces and atom overlaps \\
        \hline
    \end{tabular}
    \label{table:seach_papers_questions}
\end{table}

Complementary to PaperQA's closed-corpus retrieval, we enabled the LLMs internal web search tool, and set a medium context size. This tool provides access to up-to-date, open-web resources, such as papers, software documentation, and community-reported best practices. The medium context setting returns passages of sufficient granularity for the model to extract protocol-level details without overwhelming the reasoning steps of the agent.

Together, the web search tool and PaperQA create a hybrid retrieval environment. This combination improves the agent's ability to generalize across simulation systems, validate parameter choices, and adapt to evolving methods in the MD community. 

\subsection{Analysis and Visualization}
Upon successful completion of the MD run, the analyzer performs automated trajectory analysis.
It generates RMSD, radius of gyration, RMSF, and hydrogen-bond statistics, using GROMACS analysis tools.
Each plot is accompanied by an automatically generated textual interpretation describing structural stability, folding dynamics, and binding behaviour.

\section{Code availability}
The code to reproduce results and utilize the agentic framework with open-source softwares is available at \url{https://github.com/schwallergroup/DynaMate}.

\section{Authors contributions}
Using the CRediT system: Conceptualization: CM, SG, JG, \& PS; Data curation: CM, SG \& JG; Formal analysis: CM, SG \& JG; Investigation: CM, SG, JG \& PS; Methodology: CM, SG, JG \& BN. 
Project administration: PS; Resources: PS; Software: CM, SG JG \& BN; Supervision: PS; Validation: CM, SG \& JG; 
Visualization: CM, SG \& JG; Writing -- original draft: CM, SG, JG, BN \& PS; Writing -- review and editing: CM, SG, JG, BN \& PS.

\section{Acknowledgments}
CM acknowledges Valence Labs for financial support. SG acknowleges support from Intel and Merck KGaA via the AWASES programme, and the Swiss National Science Foundation (SNSF) (grant number: 200020-219440). BN acknowledges funding from Swiss National Science Foundation (SNSF) (grant number: 226509). PS acknowledges support from the NCCR Catalysis (grant number 225147), a National Centre of Competence in Research funded by the Swiss National Science Foundation.

\clearpage

\bibliographystyle{elsarticle-num} 
\bibliography{References}

\clearpage

\section*{Supplementary Information}\label{SI: Supplementary Information}
\renewcommand{\thefigure}{S\arabic{figure}}
\setcounter{figure}{0}
\renewcommand{\thetable}{S\arabic{table}}
\setcounter{table}{0}
\setcounter{section}{0}
\renewcommand{\thesubsection}{\Alph{subsection}}

\subsection{\textbf{\agentname~capabilities}}
Here, we outline the current state of both existing agentic frameworks that run MD simulations, and our framework. We specify both the types of systems it can accept, and the range of tasks it can perform.
\begin{table}[ht]
    \centering
    \small
    \setlength{\tabcolsep}{5pt}
    \caption{Comparison of capabilities across existing agentic MD frameworks. \agentname~extends prior systems by enabling ligand handling, adaptive error correction, and retrieval-augmented parameter selection.}
    \label{SI:framework_comparison}
    \vspace{0.3em}
    \begin{tabularx}{\linewidth}{l *{3}{>{\centering\arraybackslash}X}}
        \toprule
        \textbf{Capability} & \textbf{MDCrow} & \textbf{NAMD-Agent} & \textbf{\agentname~(ours)} \\
        \midrule
        Protein--only simulations & Yes & Yes & Yes \\
        Protein--ligand simulations & No & No & Yes \\
        Protein--ligand binding affinity calculations & No & No & Yes \\
        Automated structure cleaning & Partial & Yes & Yes \\
        Force field parameterization & Yes (OpenMM) & Yes (CHARMM-GUI) & Yes (AmberTools/GROMACS) \\
        Retrieval from literature/databases & Yes (limited) & No & Yes (RAG via RCSB/PubChem/Google) \\
        Adaptive tool selection & No & Partial & Yes \\
        Error correction and recovery & No & No & Yes \\
        Natural-language task specification & Yes & Yes & Yes \\
        Modular multi-agent architecture & No & Partial & Yes \\
        Cross-platform compatibility & OpenMM only & CHARMM only & GROMACS / AmberTools \\
        \bottomrule
    \end{tabularx}
\end{table}

\subsection{Human defined workflow}\label{tab:human_defined_workflow}

There is a series of tools provided to the agent, that it should use correctly for simulation success. 
The tools provided include:

\begin{itemize}
    \item Fetch and save the pdb file.
    \item Preparation of pdb files. This comprises cleaning the protein file, capping it (C- and N- termini with ACE and NME), extracting the ligand file, and protonating it at pH = 7 using Open Babel~\cite{o2011open}. 
    \item Preparation of ligand parameters with AmberTools.
    \item Generation of the protein-ligand complex file if a ligand is present.
    \item Web and data retrieval: query of external sources (RCSB, PubChem, Google) to determine experimental conditions such as recommended temperature or ligand information.
    \item Generation parameters for the protein or protein-ligand systems using tLEaP. The protein is parameterized with the ff14sb force field and the ligand with GAFF2, with AM1-BCC charges.
    \item Preparation the simulation box with AmberTools, by solvating with the TIP3P model and neutralising with ions.
    \item Equilibration the system with GROMACS: it includes energy minimisation until the maximum force on all atoms is inferior to 10.0 kJ/mol, 100 ps NVT and 100 ps NPT simulations to equilibrate the temperature and pressure of the system. Both properties are plotted and analyzed for convergence.
    \item Generation of the NPT production run with GROMACS, with the temperature either specified by the user, or determined by the agent with its web-search tool.
    \item Analysis of the RMSD, RMSF, radius of gyration, number of hydrogen-bonds (between protein side-chains, protein-ligand and protein-water) plots generated by GROMACS.
\end{itemize}

\begin{figure}[th]
    \centering
    \includegraphics[width=0.8\linewidth]{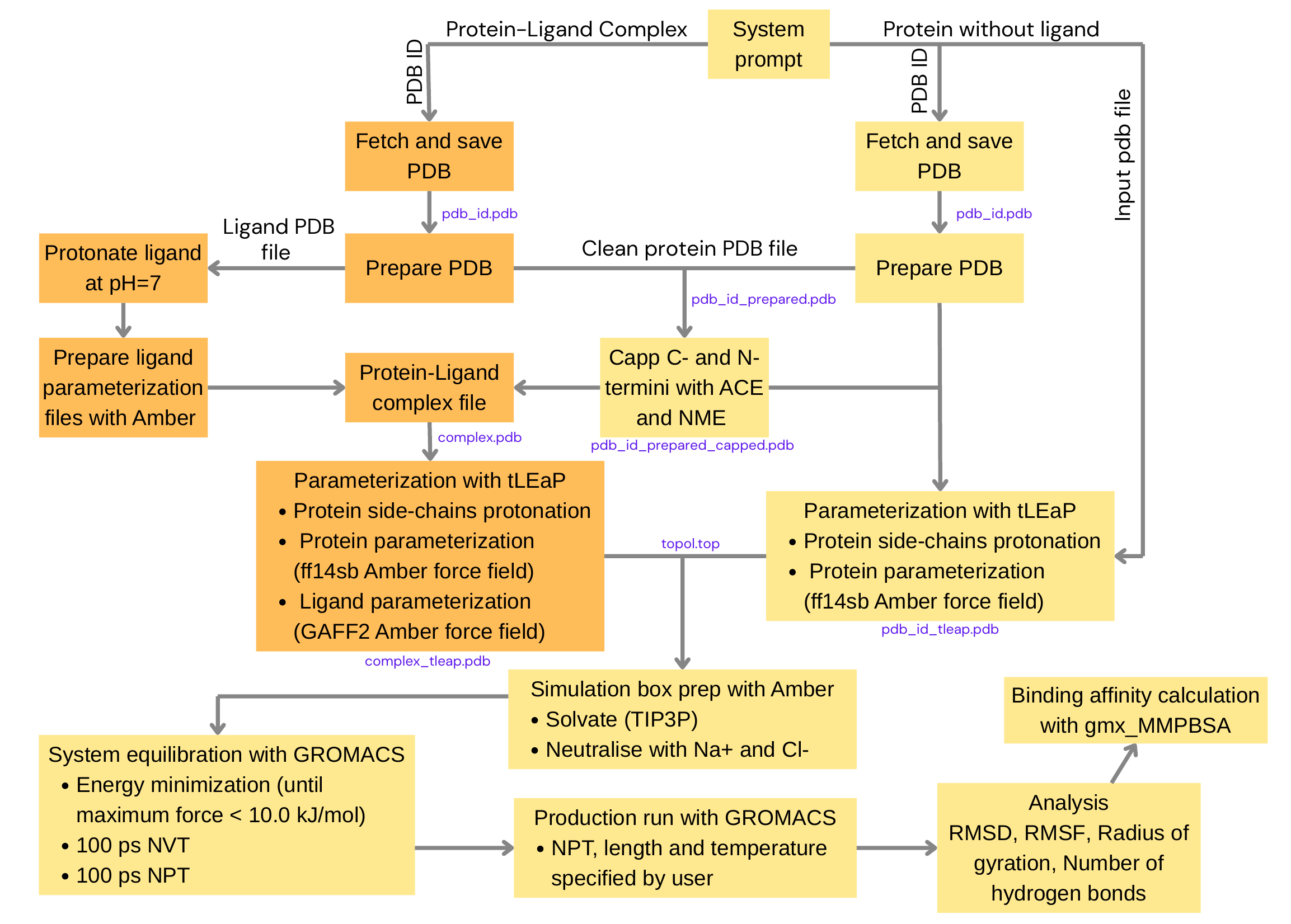}
    \caption{Agentic workflow for a protein-only system (yellow) and a protein-ligand system (orange+yellow). Steps include structure retrieval, preprocessing, generation of input force field parameters, solvation, equilibration, and production.}
    \label{fig:workflow_tools_protein_ligand}
\end{figure}


\subsection{Reference systems for evaluation}

\begin{table}[htbp]
\centering
\caption{Systems used to evaluate the LLM agent for molecular dynamics simulations (Protein-ligand complexes and Protein systems).}
\label{SI:systems_choice_rationale}
\small
\begin{tabularx}{\textwidth}{l p{0.22\textwidth} l X}
\toprule
\textbf{PDB ID} & \textbf{Protein} & \textbf{Ligand} & \textbf{Rationale for Selection} \\
\midrule
1AKI~\cite{artymiuk1982structures} & Lysozyme & / & High resolution crystal structure of a protein monomer. \\
1FDH~\cite{frier1977structure} & Haemoglobin & / & High resolution crystal structure of a protein multimer. Good case study for multiple chain system. \\
1J37~\cite{kim2003crystal} & Drosophila AnCE & / & High resolution crystal structure of a protein monomer. \\
2CBA~\cite{haakansson1992structure} & Carbonic anhydrase II & / & High resolution crystal structure of a protein monomer. \\
2VVB~\cite{sjoblom2009structural} & Carbonic anhydrase II & / & High resolution crystal structure of a protein monomer. \\
3HTB~\cite{boyce2009predicting} & T4 Lysozyme L99A/M102Q & JZ4 & Common model for studying polar cavity binding; probes hydrogen bonding and solvation effects. \\
3PTB~\cite{marquart1983geometry} & $\beta$-Trypsin & BEN & Canonical benchmark system for protein--ligand MD; simple and well-characterized binding site. \\
4GIH~\cite{liang2013lead} & Tyk2 (JH1 domain) & 0X5 & Popular benchmark for free energy methods; halogenated, conformationally complex inhibitor. \\
4W52~\cite{merski2015homologous} & T4 Lysozyme L99A & BNZ & Minimal hydrophobic binding model; widely used to test force fields and thermodynamics. \\
5KB6~\cite{Oliveira2017} & adenosine kinase & / & High resolution crystal structure of a well-studied protein. \\
5UEZ~\cite{wang2017fragment} & BRD4 (Bromodomain- containing protein 4) & 89G & Pharmacologically relevant target; tests handling of flexible, aromatic ligands. \\
6JJ3~\cite{jiang2019discovery} & Bromodomain-containing protein 4 (BRD4 BD1) & UNL & Modified ligand atom name to test the agent's ability to correct errors.\\
\bottomrule
\end{tabularx}
\end{table}

Bromodomain-containing protein 4 (BRD4 BD1).
Ten inhibitors (compounds \textbf{16} to \textbf{25} from Jiang et al.~\cite{jiang2019discovery}) were first docked into BRD4 BD1 (PDB: 6JJ3~\cite{jiang2019discovery}) with the software GNINA 1.3~\cite{mcnutt2025gnina}
\subsection{Extended error correction ability and model efficiency}\label{SI:sec:extended_tool}

\begin{figure}[H]
    \centering
    \def\svgwidth{\textwidth}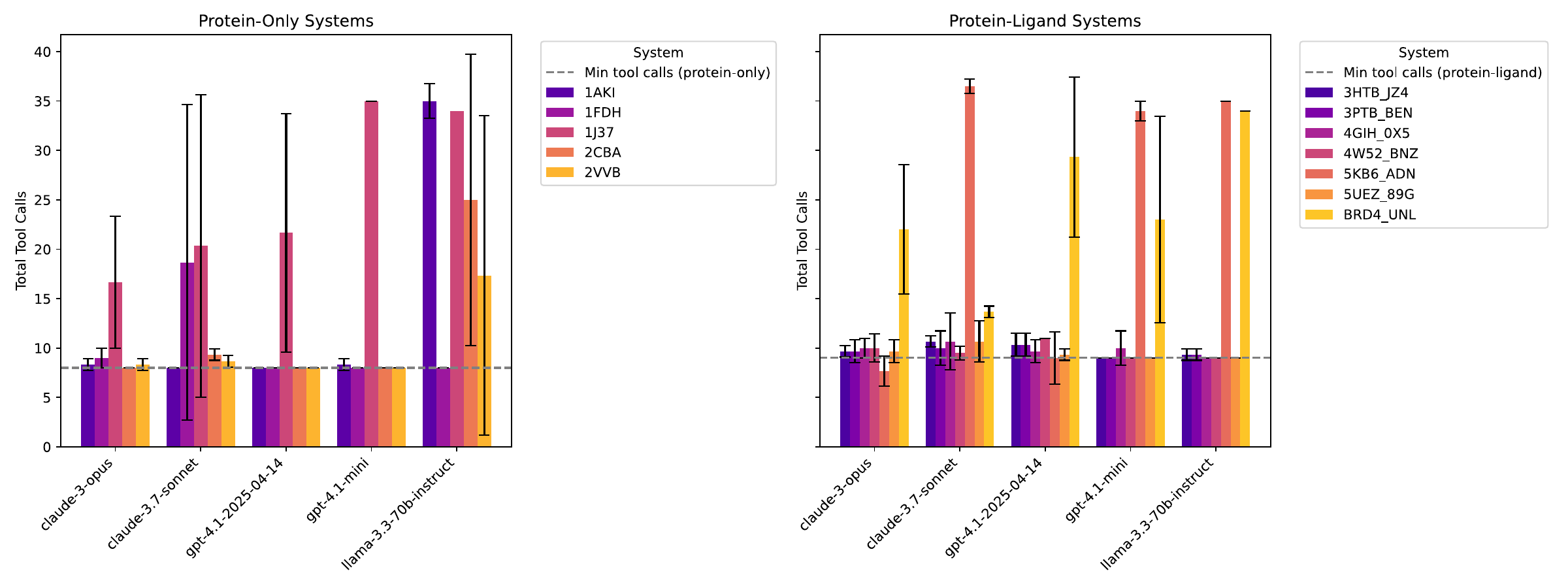
    \caption{The number of tool calls by each model separated into protein-ligand and protein-only systems. Again we denote the minimum number of tools calls by the horizontal dashed lines.}
    \label{app:fig:tool_calls_by_system}
\end{figure}

\begin{figure}[H]
    \centering
    \def\svgwidth{\textwidth}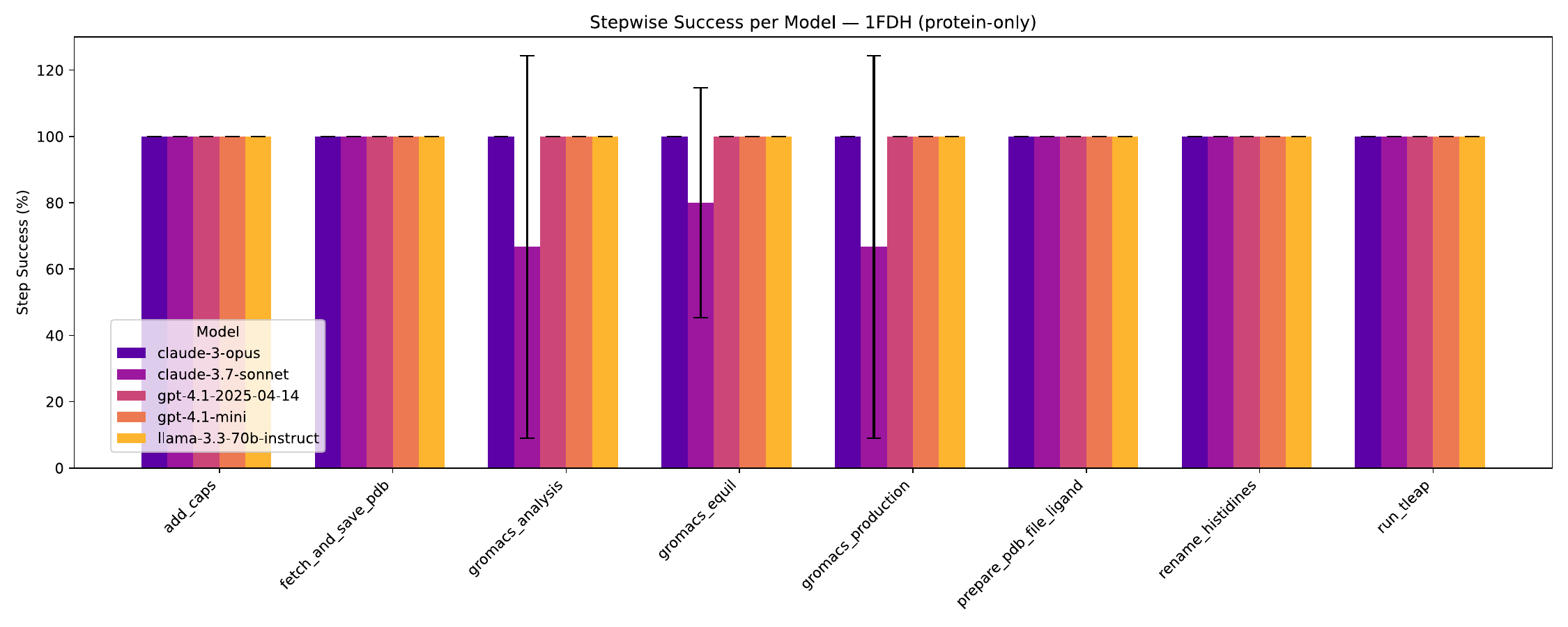
    
    \caption{The overall success of the pipeline can be broken down into the success rate of each required step. Here we show the breakdown of the success for each step for the different LLMs on the system 1FDH. We test each system three times.}
    \label{app:fig:problem_systems_1FDH}
\end{figure}

\begin{figure}[H]
    \centering
    \def\svgwidth{\textwidth}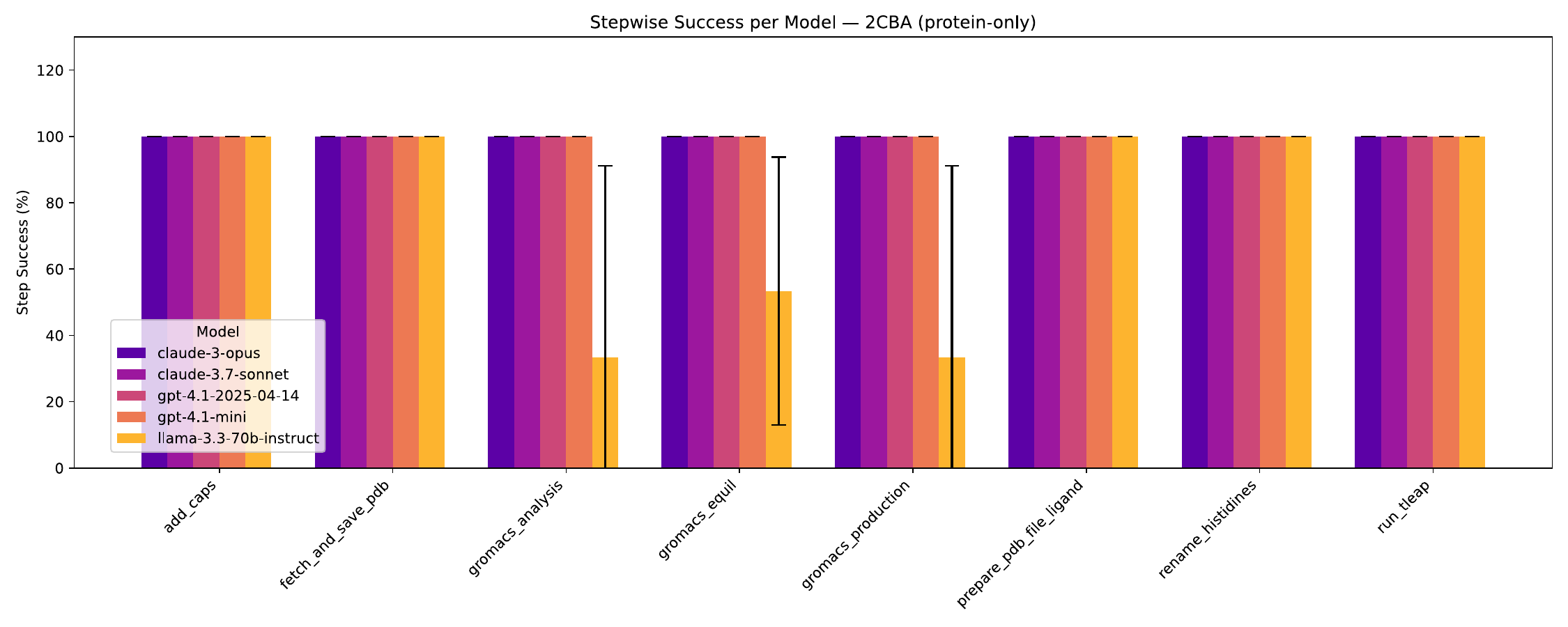
    \caption{The overall success of the pipeline can be broken down into the success rate of each required step. Here we show the breakdown of the success for each step for the different LLMs on the system 2CBA. We test each system three times.}
    \label{app:fig:problem_systems_2CBA}
\end{figure}

\begin{figure}[H]
    \centering
    \def\svgwidth{\textwidth}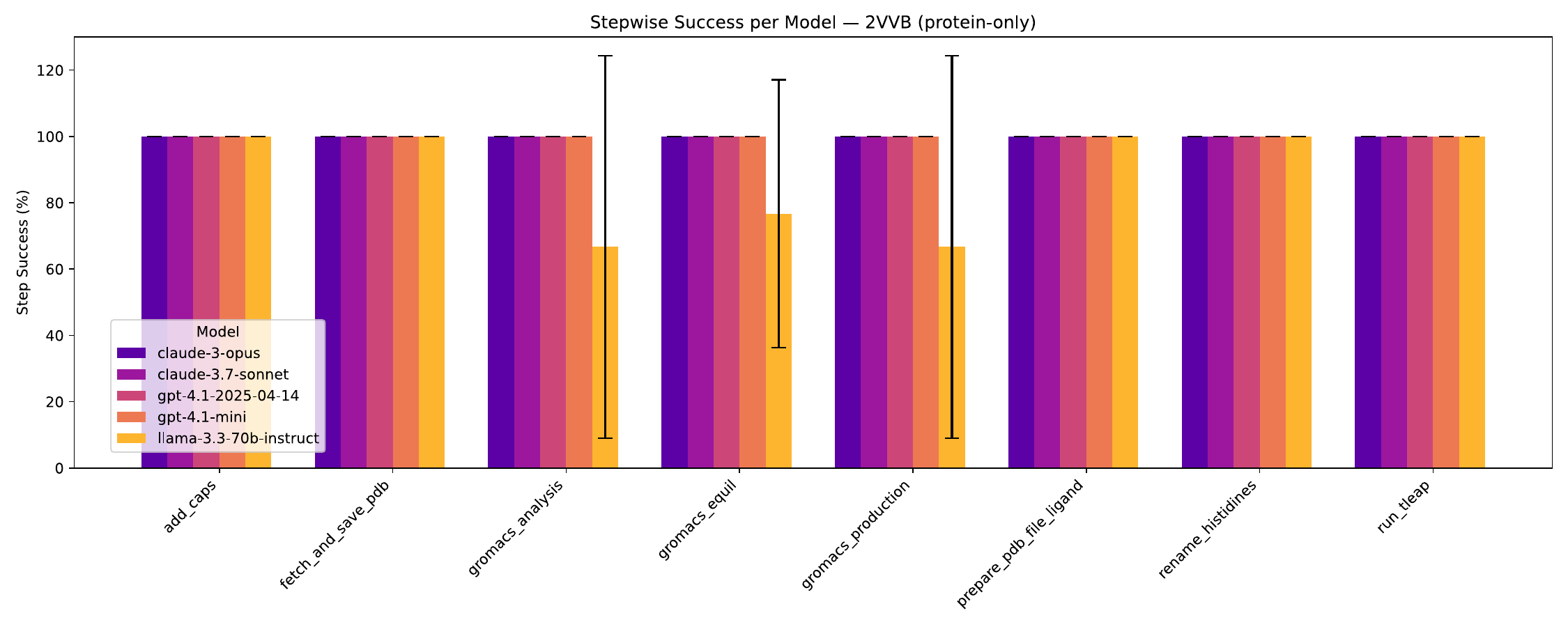
    
    \caption{The overall success of the pipeline can be broken down into the success rate of each required step. Here we show the breakdown of the success for each step for the different LLMs on the system 2VVB. We test each system three times.}
    \label{app:fig:problem_systems_2VVB}
\end{figure}

\subsection{Specialized system prompts}
\label{SI: system_prompts}
\begin{lstlisting}
PREP_SYSTEM_PROMPT = 
You are a helpful science assistant designed to fetch information about protein
structures and ligands, and make helpful suggestions regarding molecular systems.
Classify the user request and prepare the input files for an appropriate molecular
dynamics pipeline. Depending on the user inputs you should define what a successful
MD pipeline would involve. Call the relevant tools when needed to prepare the system
for molecular dynamics.
\end{lstlisting}

\begin{lstlisting}
MD_SYSTEM_PROMPT = 
You are an MD execution assistant. You have access to tools that prepare and
run molecular dynamics (MD) simulations using GROMACS.

The user has already provided a PDB structure file at {pdb_path} and the
necessary MDP files to run GROMACS. They are available in the sandbox directory
located at {sandbox_dir}, and you should use them.

You should use the tools to solvate, equilibrate, and run MD in sequence,
starting with a simulation using GROMACS and the Amber force field ff14sb.

First, if the system is a protein-ligand complex, separate the protein and
ligands into two separate pdb files. If a ligand is present, protonate it at the
appropriate pH. Check that the protonation is accurate based on a literature search.
Then, if a ligand is present, parameterise the ligand using antechamber with Amber.
If the system is a protein-ligand complex, merge the two into complex.pdb,
otherwise if the system is a protein alone, keep only the protein atoms.

Next, parameterise the system using tleap, which allows to create a box,
solvate, add ions to neutralise and prepare the 'topol.top' file. In the tleap step,
the protein will be protonated. Check that the protonation is accurate by doing
a literature search, and that the protein was protonated at the optimum pH.

Then, perform a short energy minimization using GROMACS.
Next, equilibrate with short NVT and NPT runs.
Finally, perform the production run.

After production run is complete, perform a basic analysis of the trajectory
including RMSD, RMSF calculations, radius of gyration, and hydrogen bond analysis.
The analysis of these plots should be saved as a text file named "analysis.txt"
in the sandbox directory.

If any step fails, retry after analyzing the provided error message and make
corrections to the inputs for the current step.
\end{lstlisting}

\subsection{LLM specifications}
\label{SI: models}

\begin{table}[h]
\centering
\caption{Overview of evaluated large language models and evaluation settings.}
\label{SI:tab_llm-overview}

\begin{tabularx}{\textwidth}{l l X c c c c}
\toprule
\textbf{Provider} & \textbf{Model} & \textbf{Identifier} &
\textbf{Tools} & \textbf{Web} & \textbf{Temp} & \textbf{WS Ctx} \\
\midrule

Anthropic & Claude 3 Opus &
anthropic/claude-3-opus &
Y & Y & 0.1 & Medium \\

Anthropic & Claude 3.7 Sonnet &
anthropic/claude-3.7-sonnet &
Y & Y & 0.1 & Medium \\

OpenAI & GPT-4.1 (2025-04-14) &
openai/gpt-4.1-2025-04-14 &
Y & Y & 0.1 & Medium \\

OpenAI & GPT-4.1 Mini &
openai/gpt-4.1-mini &
Y & Y & 0.1 & Medium \\

Meta & Llama 3.3 70B Instruct &
meta-llama/llama-3.3-70b-instruct &
Y & Y & 0.1 & Medium \\
\bottomrule
\end{tabularx}
\end{table}

\clearpage
\subsection{Examples of PaperQA answers} \label{SI:seach_papers_answers}

\begin{table}[ht]
    \centering
    \small
    \caption{Examples of answers generated by PaperQA following the agent questions using the {\ttfamily search\_papers} tool.}
    \begin{tabular}{|p{\textwidth}|}
        \hline
        \textbf{Question: molecular dynamics simulation temperature for 3PTB protein} \\
        \hline
        The molecular dynamics (MD) simulation temperature for the 3PTB protein is explicitly stated as 298.15 K. This temperature is achieved during the heating phase, where the system is gradually heated from 10 K to 298.15 K over 100 ps at constant volume using a Langevin thermostat with a collision frequency of 1.0 ps$^{-1}$. Subsequent simulations are conducted at this constant temperature and pressure (heinzelmann2021automationofabsolute pages 10-11). Other excerpts discuss general MD methodologies, including heating, equilibration, and production phases, but do not provide specific temperature details for the 3PTB protein (patodia2014moleculardynamicssimulation pages 2-3, patodia2014moleculardynamicssimulation pages 1-1, patodia2014moleculardynamicssimulation pages 3-4). Additionally, while one source describes simulations conducted at temperatures ranging from 300 K to 320 K, this information pertains to a different system and is not relevant to the 3PTB protein (lamprakis2021evaluatingtheefficiency pages 38-41). Thus, the only directly relevant information specifies that the MD simulation temperature for the 3PTB protein is 298.15 K (heinzelmann2021automationofabsolute pages 10-11). \\
        \hline\hline
        \textbf{Question: Why does ParmEd fail with a FileNotFoundError when using run\_tleap?}
        \\
        \hline
        ParmEd may fail with a FileNotFoundError when using run\_tleap due to missing or improperly referenced template or parameter files required for the workflow. The AMBER parmbsc0 force field, commonly used for DNA simulations, relies on templates generated using tools like TINKER, which define parameters such as charges, bonds, and angles for residues in specific chain positions. If these template files are absent or incorrectly specified, errors like FileNotFoundError can occur during the execution of run\_tleap (lopez2015mappingbiophysicsthrough pages 49-51, lopez2015mappingbiophysicsthrough pages 51-54). Additionally, topology files must be consistent and properly parameterized for tools like ParmEd and run\_tleap to function correctly. Inconsistencies or missing files, such as topology or input files, can lead to such errors. The Python utility ante-MMPBSA.py can assist in generating consistent topology files, which may help prevent these issues (miller2012<i>mmpbsa.py<i>anefficient pages 3-4). Errors may also arise from missing or misconfigured input files, such as PQR files or solvent model parameters, which are critical for molecular dynamics workflows. Proper configuration and verification of these files are essential to avoid FileNotFoundError during simulations (lopez2015mappingbiophysicsthrough pages 56-58). In summary, the failure is likely due to missing or improperly referenced template, parameter, or input files critical for the workflow (pqac-e03c3d8a, miller2012<i>mmpbsa.py<i>anefficient pages 3-4, lopez2015mappingbiophysicsthrough pages 56-58). \\
        \hline\hline
        \textbf{Question: Common issues with GROMACS energy minimization infinite forces and atom overlaps} \\
        \hline
        Common issues with energy minimization in GROMACS, such as infinite forces and atom overlaps, often arise from poorly prepared initial configurations or suboptimal parameter settings. Infinite forces typically result from atom overlaps or unrealistic geometries in the starting structure, which can occur due to improper solvation, ion placement, or packing of solvent molecules in the simulation box (balintUnknownyearsystematicexplorationof pages 13-15, lemkul2024introductorytutorialsfor pages 3-4). These issues can also stem from incorrect topology files or inadequate minimization parameters, such as improperly defined nonbonded cutoffs or force field settings (balintUnknownyearsystematicexplorationof pages 103-105, lemkul2024introductorytutorialsfor pages 4-6). To mitigate these problems, careful preparation of the system is essential. This includes ensuring proper solvation, avoiding steric clashes, and validating input files such as topology and restraint files (balintUnknownyearsystematicexplorationof pages 103-105, lemkul2024introductorytutorialsfor pages 4-6). Applying position restraints during energy minimization can help stabilize the system and prevent unphysical movements, particularly for complex systems or those with structural ions (balintUnknownyearsystematicexplorationof pages 103-105, andrio2019bioexcelbuildingblocks pages 5-7). Gradual heating during equilibration and the use of robust algorithms, such as the steepest-descent method and the LINCS algorithm for bond constraints, further reduce the likelihood of infinite forces and overlaps (balintUnknownyearsystematicexplorationof pages 13-15, andrio2019bioexcelbuildingblocks pages 5-7). Properly setting physically sound parameters in the .mdp file, such as nonbonded cutoffs consistent with the chosen force field, is critical to achieving a stable energy minimization process (lemkul2024introductorytutorialsfor pages 4-6). \\
        \hline
    \end{tabular}
\end{table}

\end{document}

%% file: figures_svg/final_heatmap_plasma_orange_blue_svg-tex.pdf_tex
\begingroup%
  \makeatletter%
  \providecommand\color[2][]{%
    \errmessage{(Inkscape) Color is used for the text in Inkscape, but the package 'color.sty' is not loaded}%
    \renewcommand\color[2][]{}%
  }%
  \providecommand\transparent[1]{%
    \errmessage{(Inkscape) Transparency is used (non-zero) for the text in Inkscape, but the package 'transparent.sty' is not loaded}%
    \renewcommand\transparent[1]{}%
  }%
  \providecommand\rotatebox[2]{#2}%
  \newcommand*\fsize{\dimexpr\f@size pt\relax}%
  \newcommand*\lineheight[1]{\fontsize{\fsize}{#1\fsize}\selectfont}%
  \ifx\svgwidth\undefined%
    \setlength{\unitlength}{1008bp}%
    \ifx\svgscale\undefined%
      \relax%
    \else%
      \setlength{\unitlength}{\unitlength * \real{\svgscale}}%
    \fi%
  \else%
    \setlength{\unitlength}{\svgwidth}%
  \fi%
  \global\let\svgwidth\undefined%
  \global\let\svgscale\undefined%
  \makeatother%
  \begin{picture}(1,0.42857143)%
    \lineheight{1}%
    \setlength\tabcolsep{0pt}%
    \put(0,0){\includegraphics[width=\unitlength,page=1]{figures_svg/final_heatmap_plasma_orange_blue_svg-tex.pdf}}%
  \end{picture}%
\endgroup%

%% file: figures_svg/final_tool_calls_svg-tex.pdf_tex
\begingroup%
  \makeatletter%
  \providecommand\color[2][]{%
    \errmessage{(Inkscape) Color is used for the text in Inkscape, but the package 'color.sty' is not loaded}%
    \renewcommand\color[2][]{}%
  }%
  \providecommand\transparent[1]{%
    \errmessage{(Inkscape) Transparency is used (non-zero) for the text in Inkscape, but the package 'transparent.sty' is not loaded}%
    \renewcommand\transparent[1]{}%
  }%
  \providecommand\rotatebox[2]{#2}%
  \newcommand*\fsize{\dimexpr\f@size pt\relax}%
  \newcommand*\lineheight[1]{\fontsize{\fsize}{#1\fsize}\selectfont}%
  \ifx\svgwidth\undefined%
    \setlength{\unitlength}{720bp}%
    \ifx\svgscale\undefined%
      \relax%
    \else%
      \setlength{\unitlength}{\unitlength * \real{\svgscale}}%
    \fi%
  \else%
    \setlength{\unitlength}{\svgwidth}%
  \fi%
  \global\let\svgwidth\undefined%
  \global\let\svgscale\undefined%
  \makeatother%
  \begin{picture}(1,0.6)%
    \lineheight{1}%
    \setlength\tabcolsep{0pt}%
    \put(0,0){\includegraphics[width=\unitlength,page=1]{figures_svg/final_tool_calls_svg-tex.pdf}}%
  \end{picture}%
\endgroup%

%% file: figures_svg/final_combined_stepwise_success_svg-tex.pdf_tex
\begingroup%
  \makeatletter%
  \providecommand\color[2][]{%
    \errmessage{(Inkscape) Color is used for the text in Inkscape, but the package 'color.sty' is not loaded}%
    \renewcommand\color[2][]{}%
  }%
  \providecommand\transparent[1]{%
    \errmessage{(Inkscape) Transparency is used (non-zero) for the text in Inkscape, but the package 'transparent.sty' is not loaded}%
    \renewcommand\transparent[1]{}%
  }%
  \providecommand\rotatebox[2]{#2}%
  \newcommand*\fsize{\dimexpr\f@size pt\relax}%
  \newcommand*\lineheight[1]{\fontsize{\fsize}{#1\fsize}\selectfont}%
  \ifx\svgwidth\undefined%
    \setlength{\unitlength}{1296bp}%
    \ifx\svgscale\undefined%
      \relax%
    \else%
      \setlength{\unitlength}{\unitlength * \real{\svgscale}}%
    \fi%
  \else%
    \setlength{\unitlength}{\svgwidth}%
  \fi%
  \global\let\svgwidth\undefined%
  \global\let\svgscale\undefined%
  \makeatother%
  \begin{picture}(1,0.66666667)%
    \lineheight{1}%
    \setlength\tabcolsep{0pt}%
    \put(0,0){\includegraphics[width=\unitlength,page=1]{figures_svg/final_combined_stepwise_success_svg-tex.pdf}}%
  \end{picture}%
\endgroup%

%% file: figures_svg/two_panel_tool_calls_svg-tex.pdf_tex
\begingroup%
  \makeatletter%
  \providecommand\color[2][]{%
    \errmessage{(Inkscape) Color is used for the text in Inkscape, but the package 'color.sty' is not loaded}%
    \renewcommand\color[2][]{}%
  }%
  \providecommand\transparent[1]{%
    \errmessage{(Inkscape) Transparency is used (non-zero) for the text in Inkscape, but the package 'transparent.sty' is not loaded}%
    \renewcommand\transparent[1]{}%
  }%
  \providecommand\rotatebox[2]{#2}%
  \newcommand*\fsize{\dimexpr\f@size pt\relax}%
  \newcommand*\lineheight[1]{\fontsize{\fsize}{#1\fsize}\selectfont}%
  \ifx\svgwidth\undefined%
    \setlength{\unitlength}{1152bp}%
    \ifx\svgscale\undefined%
      \relax%
    \else%
      \setlength{\unitlength}{\unitlength * \real{\svgscale}}%
    \fi%
  \else%
    \setlength{\unitlength}{\svgwidth}%
  \fi%
  \global\let\svgwidth\undefined%
  \global\let\svgscale\undefined%
  \makeatother%
  \begin{picture}(1,0.375)%
    \lineheight{1}%
    \setlength\tabcolsep{0pt}%
    \put(0,0){\includegraphics[width=\unitlength,page=1]{figures_svg/two_panel_tool_calls_svg-tex.pdf}}%
  \end{picture}%
\endgroup%

%% file: figures_svg/stepwise_success_1FDH_svg-tex.pdf_tex
\begingroup%
  \makeatletter%
  \providecommand\color[2][]{%
    \errmessage{(Inkscape) Color is used for the text in Inkscape, but the package 'color.sty' is not loaded}%
    \renewcommand\color[2][]{}%
  }%
  \providecommand\transparent[1]{%
    \errmessage{(Inkscape) Transparency is used (non-zero) for the text in Inkscape, but the package 'transparent.sty' is not loaded}%
    \renewcommand\transparent[1]{}%
  }%
  \providecommand\rotatebox[2]{#2}%
  \newcommand*\fsize{\dimexpr\f@size pt\relax}%
  \newcommand*\lineheight[1]{\fontsize{\fsize}{#1\fsize}\selectfont}%
  \ifx\svgwidth\undefined%
    \setlength{\unitlength}{1080bp}%
    \ifx\svgscale\undefined%
      \relax%
    \else%
      \setlength{\unitlength}{\unitlength * \real{\svgscale}}%
    \fi%
  \else%
    \setlength{\unitlength}{\svgwidth}%
  \fi%
  \global\let\svgwidth\undefined%
  \global\let\svgscale\undefined%
  \makeatother%
  \begin{picture}(1,0.4)%
    \lineheight{1}%
    \setlength\tabcolsep{0pt}%
    \put(0,0){\includegraphics[width=\unitlength,page=1]{figures_svg/stepwise_success_1FDH_svg-tex.pdf}}%
  \end{picture}%
\endgroup%

%% file: figures_svg/stepwise_success_2CBA_svg-tex.pdf_tex
\begingroup%
  \makeatletter%
  \providecommand\color[2][]{%
    \errmessage{(Inkscape) Color is used for the text in Inkscape, but the package 'color.sty' is not loaded}%
    \renewcommand\color[2][]{}%
  }%
  \providecommand\transparent[1]{%
    \errmessage{(Inkscape) Transparency is used (non-zero) for the text in Inkscape, but the package 'transparent.sty' is not loaded}%
    \renewcommand\transparent[1]{}%
  }%
  \providecommand\rotatebox[2]{#2}%
  \newcommand*\fsize{\dimexpr\f@size pt\relax}%
  \newcommand*\lineheight[1]{\fontsize{\fsize}{#1\fsize}\selectfont}%
  \ifx\svgwidth\undefined%
    \setlength{\unitlength}{1080bp}%
    \ifx\svgscale\undefined%
      \relax%
    \else%
      \setlength{\unitlength}{\unitlength * \real{\svgscale}}%
    \fi%
  \else%
    \setlength{\unitlength}{\svgwidth}%
  \fi%
  \global\let\svgwidth\undefined%
  \global\let\svgscale\undefined%
  \makeatother%
  \begin{picture}(1,0.4)%
    \lineheight{1}%
    \setlength\tabcolsep{0pt}%
    \put(0,0){\includegraphics[width=\unitlength,page=1]{figures_svg/stepwise_success_2CBA_svg-tex.pdf}}%
  \end{picture}%
\endgroup%

%% file: figures_svg/stepwise_success_2VVB_svg-tex.pdf_tex
\begingroup%
  \makeatletter%
  \providecommand\color[2][]{%
    \errmessage{(Inkscape) Color is used for the text in Inkscape, but the package 'color.sty' is not loaded}%
    \renewcommand\color[2][]{}%
  }%
  \providecommand\transparent[1]{%
    \errmessage{(Inkscape) Transparency is used (non-zero) for the text in Inkscape, but the package 'transparent.sty' is not loaded}%
    \renewcommand\transparent[1]{}%
  }%
  \providecommand\rotatebox[2]{#2}%
  \newcommand*\fsize{\dimexpr\f@size pt\relax}%
  \newcommand*\lineheight[1]{\fontsize{\fsize}{#1\fsize}\selectfont}%
  \ifx\svgwidth\undefined%
    \setlength{\unitlength}{1080bp}%
    \ifx\svgscale\undefined%
      \relax%
    \else%
      \setlength{\unitlength}{\unitlength * \real{\svgscale}}%
    \fi%
  \else%
    \setlength{\unitlength}{\svgwidth}%
  \fi%
  \global\let\svgwidth\undefined%
  \global\let\svgscale\undefined%
  \makeatother%
  \begin{picture}(1,0.4)%
    \lineheight{1}%
    \setlength\tabcolsep{0pt}%
    \put(0,0){\includegraphics[width=\unitlength,page=1]{figures_svg/stepwise_success_2VVB_svg-tex.pdf}}%
  \end{picture}%
\endgroup%